\documentclass[10pt,twocolumn,letterpaper]{article}

\usepackage[pagenumbers]{cvpr} 

\usepackage[english]{babel}

\usepackage{amsmath}
\usepackage{amsfonts}
\usepackage{amssymb}

\usepackage{adjustbox}
\usepackage{booktabs}

\usepackage{graphicx}
\usepackage{subcaption}  
\usepackage{float}
\usepackage{float}
\usepackage{threeparttable}
\usepackage{multirow}
\usepackage{tabularx}
\usepackage{adjustbox}
\usepackage[T1]{fontenc}

\usepackage[table]{xcolor}

\usepackage{pifont}

\definecolor{cvprblue}{rgb}{0.21,0.49,0.74}
\usepackage[subtle]{savetrees}

\usepackage[pagebackref,breaklinks,colorlinks,allcolors=cvprblue]{hyperref}

\usepackage[bottom]{footmisc}
\newcommand\blfootnote[1]{%
  \begingroup
  \renewcommand\thefootnote{}\footnote{#1}%
  \addtocounter{footnote}{-1}%
  \endgroup
}

\def\paperID{15802} 
\def\confName{CVPR}
\def\confYear{2026}

\title{GEO-Bench-2: From Performance to Capability, Rethinking Evaluation in Geospatial AI}

\author{Naomi Simumba$^\dagger$\\
IBM Research Europe\\
\and
Nils Lehmann$^\dagger$\\
Technical University Munich\\
\and
Paolo Fraccaro$^{\dagger \ddagger}$\\
IBM Research Europe\\
{\tt\small paolo.fraccaro@ibm.com}
\and
Hamed Alemohammad\\
Clark University\\
\and
Geeth De Mel\\
IBM Research Europe\\
\and
Salman Khan\\
MBZUAI\\
\and
Manil Maskey\\
NASA Impact\\
\and
Nicolas Longepe\\
ESA $\Phi$-lab\\
\and
Xiao Xiang Zhu\\
Technical University Munich\\
\and
Hannah Kerner\\
Arizona State University\\
\and
Juan Bernabe-Moreno\\
IBM Research Europe\\
\and
Alexandre Lacoste$^\ddagger$\\
ServiceNow AI Research\\
{\tt\small alexandre.lacoste@servicenow.com}
}

\begin{document}
\maketitle

\blfootnote{$^\dagger$ Equal Contribution.}
\blfootnote{$\ddagger$ Corresponding authors}

\begin{abstract}
Geospatial Foundation Models (GeoFMs) are transforming Earth Observation (EO), but evaluation lacks standardized protocols. GEO-Bench-2 addresses this with a comprehensive framework spanning classification, segmentation, regression, object detection, and instance segmentation across 19 permissively-licensed datasets. We introduce ''capability'' groups to rank models on datasets that share common characteristics (e.g., resolution, bands, temporality). This enables users to identify which models excel in each capability and determine which areas need improvement in future work. To support both fair comparison and methodological innovation, we define a prescriptive yet flexible evaluation protocol. This not only ensures consistency in benchmarking but also facilitates research into model adaptation strategies, a key and open challenge in advancing GeoFMs for downstream tasks.

Our experiments show that no single model dominates across all tasks, confirming the specificity of the choices made during architecture design and pretraining. While models pretrained on natural images (ConvNext ImageNet, DINO V3) excel on high-resolution tasks, EO-specific models (TerraMind, Prithvi, and Clay) outperform them on multispectral applications such as agriculture and disaster response. These findings demonstrate that optimal model choice depends on task requirements, data modalities, and constraints. This shows that the goal of a single GeoFM model that performs well across all tasks remains open for future research. GEO-Bench-2 enables informed, reproducible GeoFM evaluation tailored to specific use cases. Code, data, and leaderboard for GEO-Bench-2 are publicly released under a permissive license.
\end{abstract}

\section{Introduction}
Deep learning and self-supervision have transformed Earth Observation (EO), enabling analysis of vast, heterogeneous geospatial data at unprecedented scales and accuracy. Geospatial Foundation Models (GeoFMs)—large-scale architectures pre-trained on diverse data—promise to generalize across tasks, sensors, and geographies while reducing reliance on task-specific supervision. However, their development remains hampered by data complexity and the absence of standardized evaluation protocols.



\begin{table*}
\centering
  \caption{Model ranking across GEO-Bench-2 capabilities}
  \label{fig:dimension_model_rank}
  \includegraphics[width=0.9\linewidth]{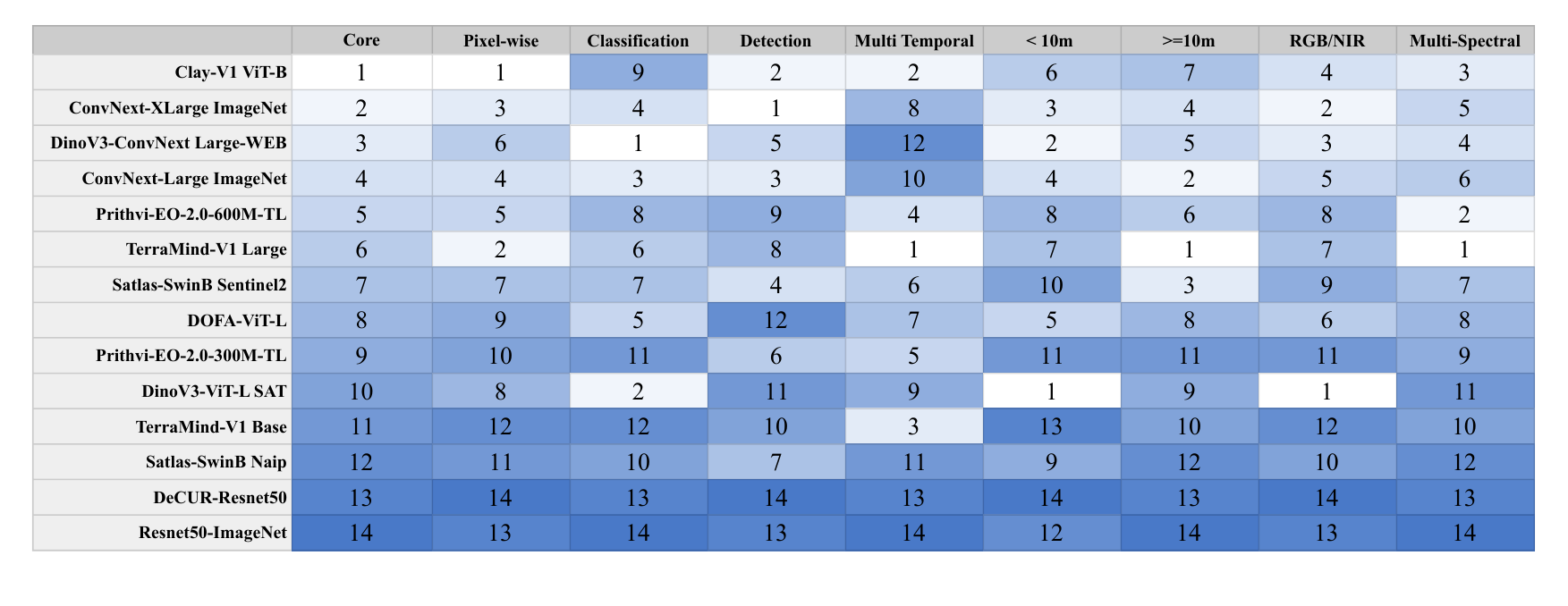}
\end{table*}

\begin{figure*}[t]
\centering
\includegraphics[width=0.8\linewidth]{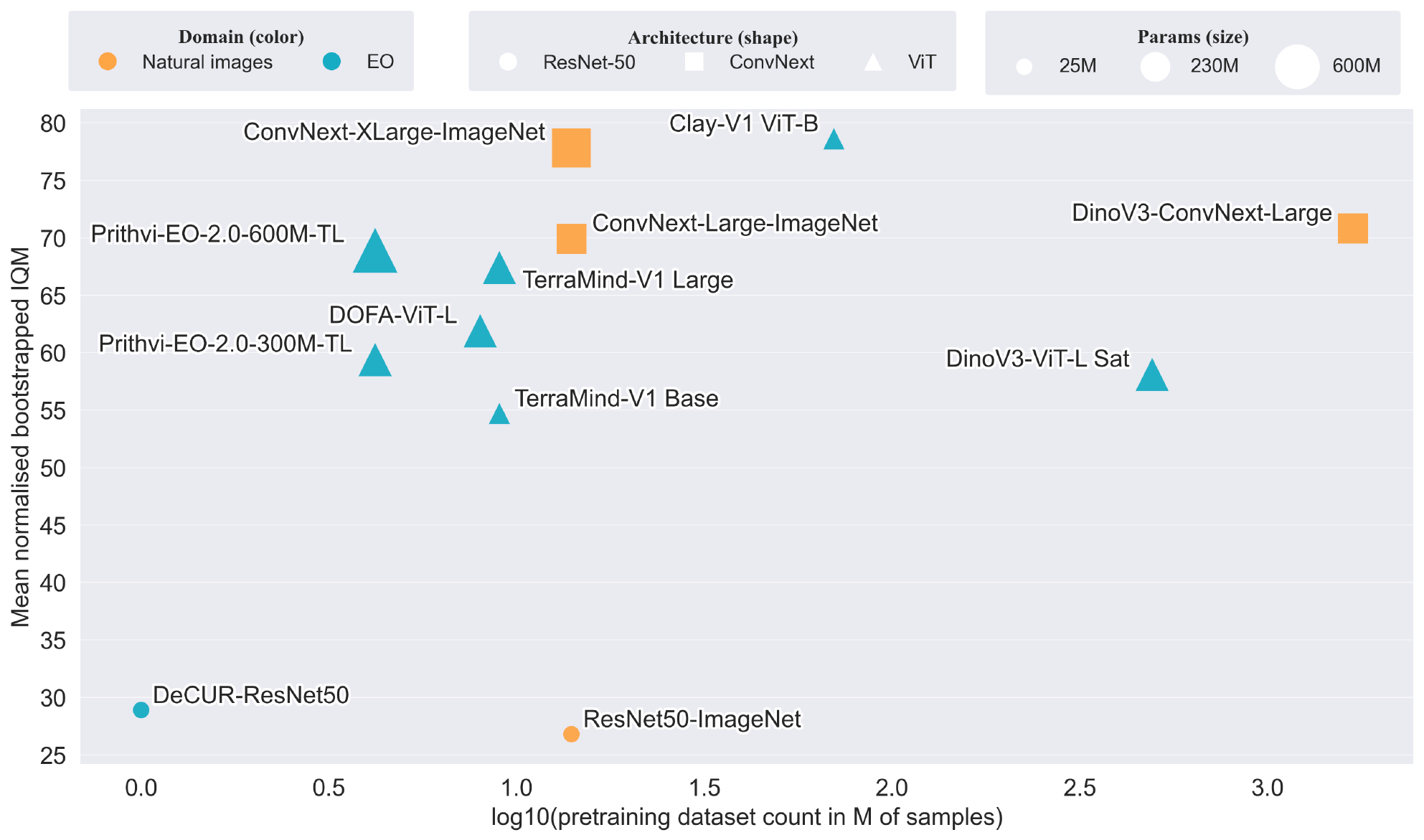}
\caption{Effect of pretraining factors on model performance on the Core capability.}
\label{fig:bubble_plot}
\end{figure*}

Benchmarking is critical in EO, where models must generalize across spatial, temporal, spectral, and sensor domains. While existing benchmarks like PANGAEA \cite{pangaea}, Copernicus-Bench \cite{copernicus_bench}, and others \cite{fomo, li2025reobench, phileo} have made valuable contributions, they remain limited by narrow modality focus, restrictive licensing, incomplete task coverage, or limited guidelines for reproducible research.
GEO-Bench \cite{geobenchv1} introduced structured evaluation protocols but had key gaps: limited task coverage (no detection or instance segmentation), lack of integration with fine-tuning tools, and inability to assess specific model capabilities beyond coarse aggregation.

We present GEO-Bench-2, a comprehensive framework for evaluating GeoFMs across real-world EO applications. Our contributions include: (1) 19 curated datasets with permissive licenses organized into overlapping capability-specific subsets (pixel-wise, detection, multi-temporal, multi-spectral, etc.); (2) geographically balanced splits for scalable, representative evaluation; (3) a prescriptive, but flexible, protocol that ensures fair comparison while allowing for innovation in model finetuning and adaptation; (4) full integration with TerraTorch \cite{terratorch}, lowering barriers to entry; and (5) an interactive leaderboard on Hugging Face for community engagement. GEO-Bench-2 enables targeted capability assessment and reproducible comparison, accelerating progress toward general-purpose geospatial intelligence.

\newcommand{\mypar}[1]{\vspace{0.5em}\noindent\textbf{#1} }

\section{GEO-Bench-2}
\subsection{Dataset Selection and Transformation}
There are hundreds of EO labeled datasets \cite{dataset_review}. Through an in-depth investigation, we identified 35 datasets with the potential to meet the criteria below. After experimentation and thorough analysis, we selected 19 that we believe enable high-quality GeoFM evaluation (see Table~\ref{tab:overview_datasets} and full details in Appendix \ref{app:datasets}). Our selection criteria were:

\mypar{Challenging and Discriminative} Datasets must clearly distinguish strong GeoFMs from baseline models. We experimentally validated discriminative power across multiple models, discarding datasets with overlapping performance profiles. For example, most well-trained models can obtain above 98\% accuracy on Eurosat \cite{eurosat} or saturate in multi-modal datasets like Sen1Floods11 \cite{sen1floods11}.

\mypar{Open Licenses} We prioritized permissive licenses to enable academic and industry adoption, avoiding GPL and non-commercial licenses\footnote{For some capabilities (e.g., detection), finding suitable open-license datasets proved challenging.}.

\begin{figure}
\centering
\includegraphics[width=1.0\columnwidth]{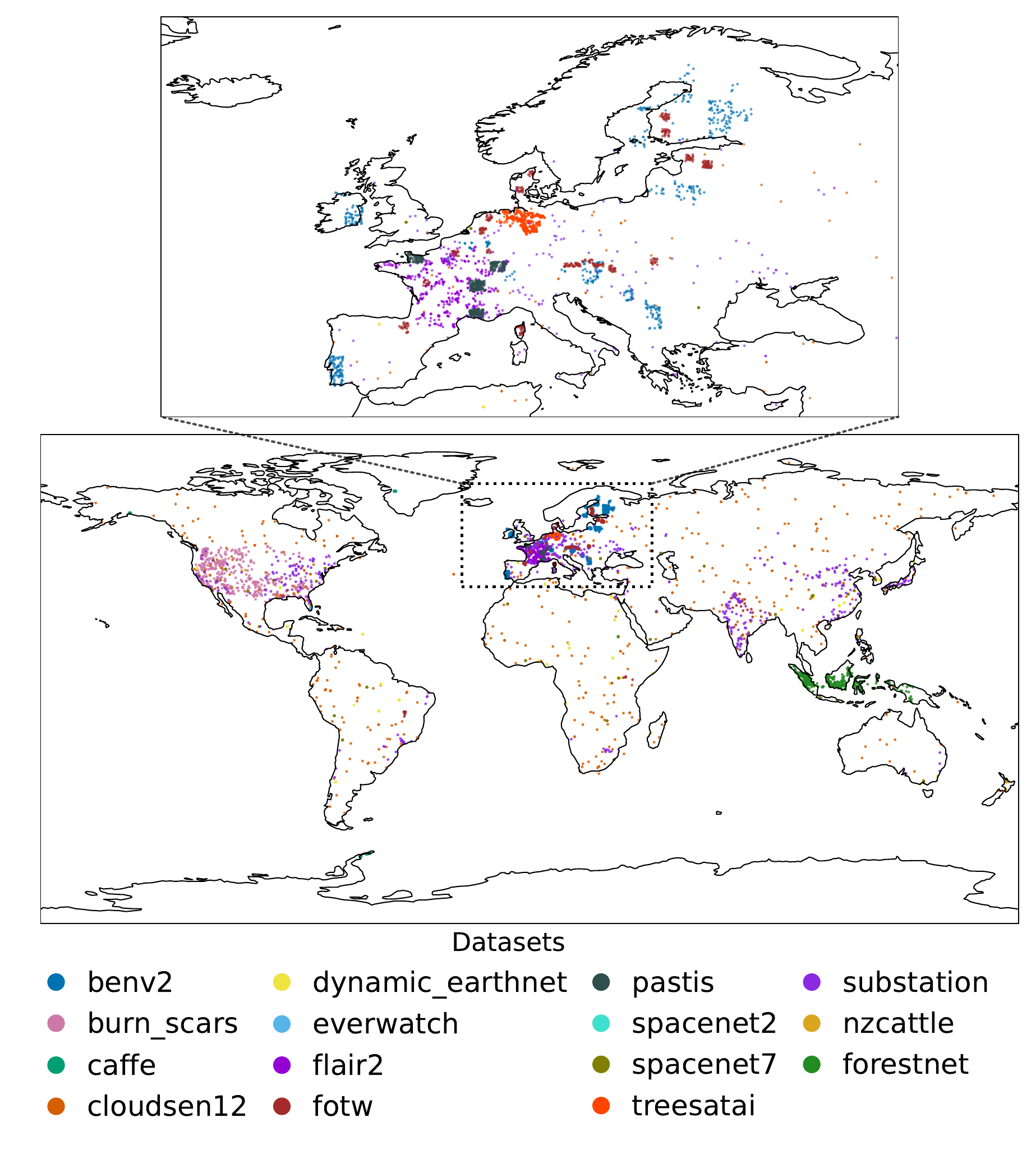}
\caption{Global distribution of samples (500 random locations per dataset shown for visualization). The BioMassters and So2Sat datasets were originally released without geospatial information and are hence not represented in this figure.}
\label{fig:geospatial_distribution}
\end{figure}

\mypar{Diversity} The benchmark encompasses a wide range of tasks, modalities, and geographic regions, featuring samples from all seven continents (Figure~\ref{fig:geospatial_distribution}). The comparatively higher representation of Europe stems from initiatives such as INSPIRE and Horizon Europe (see Figure~\ref{fig:geospatial_distribution_continets} for relative continental coverage)\footnote{To enhance global balance, we encourage the community to release open-source datasets from underrepresented regions, with an effort to avoid non-commercial restrictions. Additionally, to promote the creation of initiatives similar to INSPIRE and Horizon Europe worldwide.}.

If needed, datasets were sub-sampled to reduce computational costs. Datasets were formatted following the TACO convention \cite{aybar2025missing}, ensuring FAIR compliance \cite{wilkinson2016fair} with self-contained, ML-ready samples. If sub-sampling was applied, a maximum of 20,000 was selected for classification and 5,000 for pixel-wise and detection tasks, with the exception of the Biomasters dataset. 

\begin{table*}
\centering
\scriptsize
\rowcolors{2}{blue!3}{white}
\resizebox{\textwidth}{!}{%
\renewcommand{\arraystretch}{1.2}
\begin{tabular}{@{}l l l l c l c l@{}}
\toprule
\textbf{Task} & \textbf{Dataset} & \textbf{Domain} & \textbf{Modalities} & \textbf{GSD} & \textbf{Train/Val/Test} & \textbf{Classes} & \textbf{License} \\
\midrule
Classification 
  & BEN V2 \cite{reben} & Land cover & S1+S2 & 10m & 20000/4000/4000 & 19 & CDLA-Permissive-1.0 \\
  & TreeSatAI \cite{treesatai}  & Tree Species & S2 TS & 10m & 20000/4000/4000 & 13 & CC-BY-4.0 \\
  & So2sat \cite{so2sat, geobenchv1} & Climate Zones & S2 & 10m & 19992/986/986 & 17 & CC-BY-4.0 \\
  & Forestnet \cite{forestnet, geobenchv1} & Tree Species & L8 & 15m & 6464/989/989 & 12 & CC-BY-4.0 \\
\midrule
Pixel Regression
  & BioMassters \cite{biomassters} & Biomass estim. & S1+S2 TS & 10m & 4011/1739/2776 &  $\mathbb{R}_+$ & CC-BY-4.0 \\
\midrule
\shortstack[l]{Semantic\\Segmentation}
  & CaFFe \cite{caffe} & Glacier zones & S1 SAR & 10m & 4000/1000/2000 & 4 & CC-BY-4.0 \\
  & CloudSen12 \cite{cloudsen12+} & Cloud/shadow & S1+S2 & 10m & 4000/535/975 & 4 & CC0 \\
  & NASA Burn Scars \cite{prithvi2} & Burn scars & HLS & 30m & 524/160/120 & 2 & CC-BY-4.0 \\
  & Dynamic EarthNet \cite{dynamicearthnet} & LULC (temporal) & Planet & 3m & 700/100/200 & 7 & CC-BY-4.0 \\
  & FLAIR 2 \cite{flair2} & LULC & Aerial RGBN+DEM & 0.2m & 4049/1022/3022 & 13 & Open License 2.0 \\
  & FTW \cite{fow} & Field boundaries & S2 & 10m & 4000/1000/2000 & 2 & CC-BY-SA \\
  & KuroSiwo \cite{kurosiwo} & Flood extent & S1+DEM+Slope & 10m & 4000/1000/2000 & 4 & MIT \\
  & PASTIS (R) \cite{pastisr} & Crop type map. & S1+S2 & 10m & 1455/482/496 & 19 & CC-BY-4.0 \\
  & SpaceNet2 \cite{spacenet} & Building & Worldview & 0.3m & 5186/1461/2961  & 2 & CC-BY-SA-4.0 \\
  & SpaceNet7 \cite{spacenet} & Building & Planet & 3m & 3888/652/1152 & 2 & CC-BY-SA-4.0 \\
\midrule
Object Detection
  & EverWatch \cite{everwatch} & Bird species & Aerial RGB & 0.1m & 4429/500/196 & 9 & CC0 \\
  & m-nzcattle \cite{nzcattle} & Cattle & Aerial RGB & 0.1m & 524/66/65 & 2 & CC-BY-4.0 \\
\midrule
\shortstack[l]{Instance\\Segmentation}
  & PASTIS (R) panoptic \cite{pastisr} & Crop type map. & S1+S2 & 10m & 1455/482/496 & 19 & CC-BY-4.0 \\
  & Substations \cite{substations} & Substations & S2 & 10m & 4000/500/500 & 2 & CC-BY-4.0 \\
\bottomrule
\end{tabular}%
}
\caption{Anonymized-Bench dataset overview grouped by task category. S1: Sentinel-1; S2: Sentinel-2; DEM: Digital Elevation Model; TS: Time series; LULC: Land Use Land Cover.}
\label{tab:overview_datasets}
\end{table*}

\subsection{Capability Groups}\label{sec:capabilities}
Single-aggregate metrics obscure GeoFM's strengths and constrain research directions. GEO-Bench-2 instead evaluates nine overlapping \textbf{capabilities} designed to highlight specific challenges in GeoFM development. These include architectural challenges, such as object detection and segmentation, as well as input-related challenges like sensor fusion and temporal understanding. (Table~\ref{tab:capability_table}).

\mypar{Core} \textit{Core} provides a balanced subset across all capabilities with the aim of reducing the compute needed for evaluating on the benchmark while evaluating on the more discriminative datasets. 

\mypar{ML Task Type} \textit{Classification} includes single- and multi-label image classification tasks. \textit{Pixel-wise predictions} encompass semantic segmentation and regression, assigning classes or continuous values per pixel. \textit{Detection} evaluates object detection and instance segmentation with bounding boxes and masks. Since the detection datasets can be challenging to work with from a computational and methodological perspective, they are included only in 2 subsets to make other capabilities more accessible.

\mypar{Temporality}  \textit{Multi-temporal} datasets include multiple timestamps for change detection and temporal modeling.

\mypar{Resolution} To study over-reliance on data coming from Sentinel or Landsat, we also create subsets to study performances under and above 10 m GSD.

\mypar{Spectral Diversity} Spectrally, \textit{RGB/NIR} includes basic visible and near-infrared data, while \textit{Multi-spectral-dependent} requires optical bands beyond RGB for task completion (validated via ablation, see Appendix \ref{app:rgb_results}). We do not explicitly define a SAR capability, but five datasets (KuroSiwo, PASTIS-R, CaFFe, BEN V2, CloudSEN12) could be used to study this axis independently.


\begin{table*}[htb]
\centering
\scriptsize
\rowcolors{2}{blue!3}{white}  
\begin{tabular}{l*{9}{c}}
\toprule
\textbf{Dataset} & \textbf{Core} & \shortstack{\textbf{Pixel}\\\textbf{wise}} & \shortstack{\textbf{Classi-}\\\textbf{fication}} & \textbf{Detection} & \shortstack{\textbf{Multi}\\\textbf{Temporal}} & \shortstack{\textbf{$<10$m}\\\textbf{Res}} & \shortstack{\textbf{$\geq$10m}\\\textbf{Res}} & \shortstack{\textbf{RGB/}\\\textbf{NIR}} & \shortstack{\textbf{Multi}\\\textbf{Spectral}} \\ 
\midrule
BEN V2             & \checkmark & & \checkmark & & & & \checkmark & & \checkmark \\ 
TreeSatAI            & \checkmark & & \checkmark & & & \checkmark & & \checkmark & \\ 
So2Sat             & & & \checkmark & & & & \checkmark & & \checkmark \\ 
ForestNet          & & & \checkmark & & & & \checkmark & & \\ 
BioMassters          & \checkmark & \checkmark & & & \checkmark & & \checkmark & & \checkmark \\ 
CaFFe                & & \checkmark & & & & & \checkmark & & \\ 
CloudSEN12           & \checkmark & \checkmark & & & & & \checkmark & & \checkmark \\ 
NASA Burn Scars      & \checkmark & \checkmark & & & & & \checkmark & & \checkmark \\ 
Dynamic Earth Net     & & \checkmark & & & \checkmark & \checkmark & & \checkmark & \\ 
FLAIR 2               & \checkmark & \checkmark & & & & \checkmark & & \checkmark & \\ 
FTW               & \checkmark & \checkmark & & & & & \checkmark & \checkmark & \checkmark \\ 
KuroSiwo             & \checkmark & \checkmark & & & \checkmark & & \checkmark & & \\ 
PASTIS               & \checkmark & \checkmark & & & \checkmark & & \checkmark & & \checkmark \\ 
SpaceNet 2            & & \checkmark & & & & \checkmark & & & \checkmark \\ 
SpaceNet 7            & \checkmark & \checkmark & & & & \checkmark & & \checkmark & \\ 
EverWatch            & \checkmark & & & \checkmark & & & & & \\ 
NZCattle           & & & & \checkmark & & & & & \\ 
PASTIS (R) panoptic  & & & & \checkmark & & & & & \\ 
Substations        & \checkmark & & & \checkmark & & & & & \\ 
\bottomrule
\end{tabular}
\caption{Overview of dataset capabilities.}
\label{tab:capability_table}
\end{table*}

\newcommand{\mandatory}[1]{\textcolor{blue}{#1}}

\section{Evaluation Protocol}
The benchmark's main objective is to advance GeoFMs' development by measuring their performance on downstream tasks. However, since adapting a base model to a specific dataset remains an active research area with significant variation across models and task types, our second objective is to support this exploration. We therefore allow users flexibility in how they adapt GeoFMs to downstream tasks within our framework. Our experiments establish a \textbf{baseline adaptation protocol} and provide initial rankings for several GeoFMs, but we encourage users to submit their own leaderboard entries demonstrating improved adaptation strategies.

To ensure fair comparisons across GeoFMs, we require users to follow \mandatory{mandatory guidelines highlighted in blue throughout this section}. The remaining details describe our baseline adaptation protocol. As a first guideline, \mandatory{We request users to document precisely their adaptation protocol, following the provided template}. Also, to improve comparison, \mandatory{if a new GeoFM is evaluated, we suggest users to provide at least one entry according to the baseline protocol.}

\mypar{Why fine-tuning?} It is possible to use the benchmark with frozen backbones, and we provide a second leaderboard for users who would like to operate in this fashion. However, in Section~\ref{app:frozen_results}, we demonstrate an important performance degradation and significant change in ranking of models compared to a fine-tuning approach. For this reason, we recommend using the benchmark with a fine-tuning and hyperparameter selection approach.



\subsection{Hyperparameter Optimization}\label{sec:hparam-optimization}

\mypar{Search Space:} The search space of hyperparameters should be mostly uniform across the benchmark to prevent users from crafting a search space for each dataset. Users may design multiple search spaces for different task types (e.g., separate spaces for classification vs. segmentation), \mandatory{but must document these search spaces and the rationale behind them. Leaderboard submissions will be reviewed and may be flagged as \emph{over-engineered} if search spaces are excessively customized}. As a rule of thumb, we recommend a search space cover at least 4 datasets to ensure statistical validity of the HPO process.

\mypar{Search Budget:} To balance exploration with compute fairness, we impose a \mandatory{maximum search budget of 16 trials per dataset}, and the selection of the best configuration can only be done on the provided validation set. We allow Bayesian optimizers such as Optuna.

\mypar{Repeated Test:} Once the best configuration is obtained on the validation set, it is requested to \mandatory{repeat the fine-tuning of the best configuration 5 times} and evaluate on the test set with the required metric (see section \ref{sec:evaluation_metrics}) to account for training stochasticity.

\mypar{Baseline protocol:} Detailed configuration of our experiments is provided in Appendix~\ref{app:experimental_setup}. At a glance, we use Optuna \cite{optuna} in TerraTorch Iterate to automate the process, and we encourage users to do so when possible. The optimization is conducted over a fixed hyperparameter space across all models and all datasets, tuning Learning rate and Batch size while keeping other hyperparameters constant. Each HPO and repeated experiment iteration lasted for 50 epochs.

\subsection{Pre-Processing and Data Augmentation}

\mandatory{Pre-processing or data augmentations cannot be informed by the test set, and the methodology needs to be uniform across all datasets.}

\mypar{Baseline protocol:} We apply per-band Z-score normalization to each model\footnote{Note, this might penalize some base models expecting a different input distribution. However, we believe that the fine-tuning procedure mitigates this potential discrepancy.}, and parameters are estimated over the training split. The same normalization strategy is used for target data in regression datasets. During training, standard augmentations include horizontal and vertical flipping. For pixel-wise tasks, random cropping is applied to images larger than $224 \times 224$ pixels during training, while tiled inference is utilized for validation and testing to process the full image extent.

\subsection{Base Model Adaptation}
Users can employ different methodologies to adapt a base model for different categories of tasks, but \mandatory{a methodology should be generic and valid for at least 4 different datasets.}

\mypar{Baseline protocol:}

\mypar{\hspace{3mm}Classification:} We used a single linear layer with softmax on the encoder output.

\mypar{\hspace{3mm}Pixel-wise Tasks:} (Segmentation and Regression) We used a UNet decoder for both semantic segmentation and regression tasks, where we fed equally spaced features from the encoder's output into the UNet \cite{marti2025fine}. For transformer-based models, \textit{LearnedFeatureInterpolation}, available in TerraTorch, is applied to hierarchically structure the encoder output before passing it to the UNet \cite{unet2015}. This is a technique demonstrated to work well for ViTs compared to other choices (i.e., UPerNet) \cite{marti2025fine}.

\mypar{\hspace{3mm}Object Detection and Instance Segmentation:} We used Faster R-CNN \cite{girshick2015fast} and Mask R-CNN \cite{he2017mask}, respectively. For both tasks, a \textit{FeaturePyramidNetwork} \cite{lin2017feature} was applied, following the initial transformation of ViT features as described for pixel-wise tasks.

\subsection{Multi-Spectral Bands, SAR and Multi-Modal Datasets}
We leave the choice to users on how to process the available data sources in each dataset. Below, a description of the approach we used.

\mypar{Baseline Protocol:}

\mypar{\hspace{3mm}Multi-Spectral Bands:}All bands contained in a given dataset are utilized, provided they are compatible with the model. Where an exact match did not exist, model bands were matched to the dataset bands with the closest wavelength (i.e. matching S2 bands to WorldView).

\mypar{\hspace{3mm}Synthetic Aperture Radar:} In case a model could not handle Synthetic Aperture Radar (SAR) natively, VV and VH polarization bands were loaded as the model's RGB channels in the order VV, VH, and VV.

\mypar{\hspace{3mm}Multi-Modal Datasets:} It remains unclear how much S1 helps when S2 is available \cite{pangaea}. To investigate this, we ran an ablation (Appendix~\ref{app:multimodal_results}) on three multi-modal models (Terramind, DOFA, Clay), comparing S2-only performance to S1+S2 for multi-modal datasets. For the main results, we report the best outcome per model–dataset pair, which led to using both modalities only for Terramind on BEN V2 and BioMassters (Appendix~\ref{app:multimodal_results}); all other models use S2 only.

\subsection{Multi-Temporality}

\mypar{Baseline Protocol:} For datasets with multi-temporal inputs, each timestamp was passed through the encoder separately. The resulting encoder outputs were then averaged along the embedding dimensions before being passed to the decoder.

\subsection{Evaluation Metrics}
\label{sec:evaluation_metrics}
\mandatory{All submissions should report results using the following standardized metrics:}
\begin{itemize}
\item Semantic Segmentation: Multiclass Jaccard Index 
\item Single label Classification: Accuracy 
\item Multilabel classification: F1-score 
\item Pixel Wise Regression: Root Mean Squared Error (RMSE)
\item Instance segmentation and Object detection:  Mean Average Precision (mAP)
\end{itemize}

\subsection{Aggregated performance}

To obtain a valid aggregation and account for uncertainty introduced by random seeds, we follow the methodology introduced in \cite{geobenchv1}. It is recommended to use the leaderboard or the provided code for computing these scores based on raw results to minimize the chance of mistakes.

\mypar{Renormalization:} To report scores for a capability, we aggregate results across multiple datasets. Simple averaging can distort comparisons, as strong performance on one dataset may overshadow improvements elsewhere \cite{demvsar2006statistical}. Therefore, we normalize each dataset’s scores to [0,1] using a linear transformation with the worst and best result for a given dataset as the reference points\footnote{Similar to z-score normalization, but [0,1] improves interpretability.}. These normalization factors, computed from our fixed set of models, are published in our Anonymous-Bench-Repo along with tools for calculating final aggregated scores. All users should reuse these factors for consistency.

\mypar{Bootstrap:} 
To obtain the uncertainty over the final aggregated score, \mandatory{we use a stratified bootstrap over the collection of 5 normalized repeats for each dataset with 100 resamples}.

\mypar{IQM aggregation:} On each normalized bootstrapped iteration, \mandatory{results from each dataset in a capability are averaged across using the interquartile mean (IQM)} to discard outliers and obtain a more statistically stable mean \cite{agarwal2021deep}.

\mypar{Final Aggregation:} \mandatory{Results across the 100 normalized IQM bootstrap iterations are averaged to obtain a final score and standard deviation.}

\begin{table*}   
  \centering
  \scriptsize
  \rowcolors{2}{blue!3}{white}
  \begin{tabularx}{\textwidth}{l c c l X c c c c}
      \hline
      Model & Type & \shortstack{\# Backbone \\ Params} & \shortstack{Learning \\ Technique}  & Data & Res & N & T & License\\
      \hline
      Resnet50-ImageNet & ResNet-50 & 25M & Supervised & ImageNet-22k & NA & 14M   & 1 & Apache 2.0\\
      ConvNext-Large-ImageNet \cite{rw2019timm}         & ConvNext       & 230M & Supervised             & ImageNet-22k         & NA & 14M & 1 & Apache 2.0 \\
      ConvNext-XLarge-ImageNet  \cite{rw2019timm}        & ConvNext       & 390M & Supervised             & ImageNet-22k         & NA & 14M & 1 & Apache 2.0 \\
      DINOv3-ViT-L-SAT \cite{dinov3}         & ViT       & 300M & Distillation             & Maxar RGB         & 0.6 m & 493M & 1 & DINO V3\\
      DINOv3-ConvNext-Large-WEB \cite{dinov3}        & ConvNext       & 230M & Distillation             & LVD-1689M         & NA & 1689M & 1 & DINO V3\\
      Resnet50-DeCUR \cite{wang2024decur}    & ResNet-50 & 25M & Contrastive & Sentinel-2 & 10 m & 1M   & 1 & Apache 2.0 \\
      DOFA-ViT-300M \cite{xiong2024dofa}      & ViT       & 300M & MAE   & Sentinel-1 and -2, EnMap, Gaofen, Landsat & 1-30 m & 8M   & 1 & CC-BY-4.0 \\ 

      Clay-V1 ViT-B \cite{ClayFoundationModel}      & ViT       & 86M  & MAE   & Landsat 8 and 9, Sentinel-1 and -2, NAIP, LINZ, MODIS & 1-30 m & 70M   & 1 & Apache 2.0 \\ 
      Satlas-SwinB-Sentinel2 \cite{bastani2023satlaspretrain}\tnote{†} & Swin & 88M & Supervised         & Sentinel-2 & 10 m & NA   & 1 & ODC-BY \\
      Satlas-NAIP \cite{bastani2023satlaspretrain}\tnote{†} & Swin & 88M & Supervised         & NAIP & 1 m & NA   & 1 &  ODC-BY \\
      Prithvi-EO-V2-300M-TL  \cite{prithvi2}         & ViT       & 300M & MAE             & HLS        & 30 m & 4.2M & 4 & Apache 2.0 \\
      Prithvi-EO-V2-600M-TL  \cite{prithvi2}       & ViT       & 600M & MAE             & HLS        & 30 m & 4.2M & 4 & Apache 2.0 \\
      TerraMind-V1-Base \cite{jakubik2025terramind}        & ViT       & 86M & Correlation            & Sentinel-1 and -2, LULC, DEM, NDVI         & 10 m & 9M & 1 & Apache 2.0 \\

      TerraMind-V1-Large \cite{jakubik2025terramind}         & ViT       & 300M &       Correlation & Sentinel-1 and -2, LULC, DEM, NDVI        & 10 m & 9M & 1 & Apache 2.0 \\
      \hline
    \end{tabularx}
    \begin{tablenotes} 
    \footnotesize
      \item The pretraining dataset size (\emph{N}) is estimated where not clearly reported.
      \item Checkpoints source: Torchgeo (DOFA, DeCUR, and Satlas); TerraTorch (Prithvi and TerraMind); Timm (Resnet50-ImageNet and ConvNext-ImageNet); Meta (DINO V3).
    \end{tablenotes}
    \caption{Characteristics of the models compared: backbone type (Type), number of parameters (\# Param.), pretraining technique (Technique), pretraining data (Data), spatial resolution (Res.), number of samples (N), and number of timestamps per sample (T). Characteristics for Satlas refer to the model pretrained on Sentinel-2 downstream tasks. The pretraining sample size (N) is estimated when not explicitly reported.}
  \label{tab:GFMs_models_with_license}
\end{table*}

\subsection{Leaderboard}

To foster a community effort, we provide a leaderboard and encourage a wide range of users to submit their experiments, the good ones and the bad ones. We also encourage users to reproduce existing experiments. To lower the barrier of entry, users can submit an entry for a single capability without having to evaluate on all datasets of the benchmark. Finally, we reserve the right to reject submissions based on the following:
\begin{itemize}
    \item Does not follow the required guidelines
    \item Cannot be reproduced, e.g., closed source model
    \item Per-dataset over-engineering
    \item Any other reasons that would undermine the main objectives of this benchmark
\end{itemize}

\section{Results}\label{sec:results}



\begin{figure*}[t!]
\centering
\includegraphics[width=0.95\textwidth]{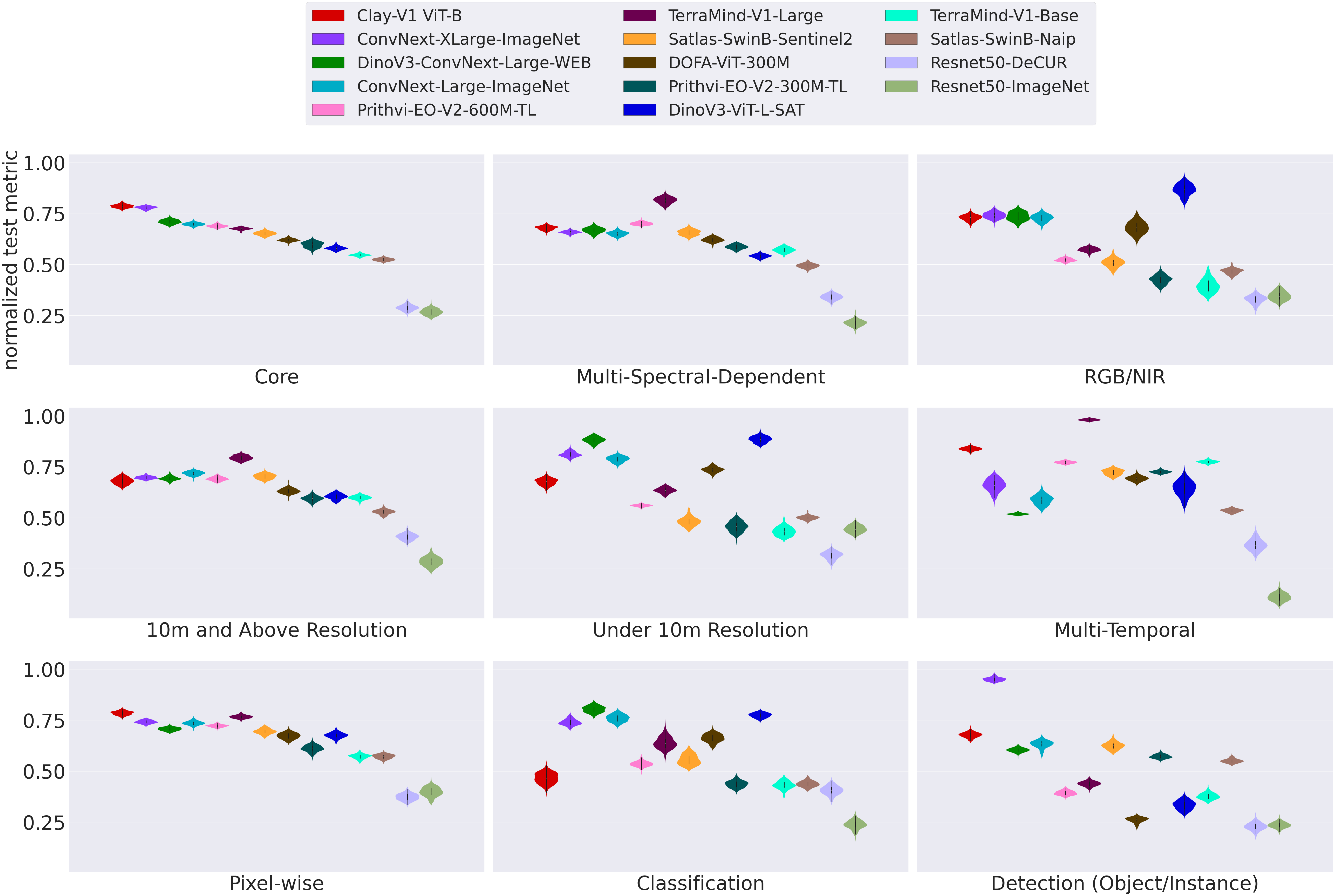}  
\caption{Normalized bootstrapped IQM performance by capability. Models are ordered based on core capability}
\label{fig:main_norm_dimensions}
\end{figure*}

We present normalized bootstrapped IQM results obtained by applying the GEO-Bench-2 protocol across all benchmarking capabilities in Figure~\ref{fig:main_norm_dimensions}. Table~\ref{fig:dimension_model_rank} provides model rankings sorted by Core capability performance. Raw performance metrics for individual datasets are available in Appendix~\ref{app:raw_results}.

\subsection{Models used in our Experiments}
To explore the value of our proposed benchmark, we have chosen models that reflect a diverse range of paradigms, model sizes, and pretraining datasets. They also include foundational general-purpose computer vision models like the latest DINOv3 \cite{dinov3} to explore their generalizability to EO tasks. The characteristics of the benchmarked models are detailed in Table~\ref{tab:GFMs_models_with_license}. All experiments in the main text are conducted with end-to-end finetuning, meaning the pretrained backbone was unfrozen during the training process.

\subsection{General Trends}
Despite some models that perform more consistently across a number of capabilities (i.e., Clay-V1 ViT-B, or ConvNeXt architectures), our results show that there is no model that dominates across all datasets. On the contrary, looking at Figure \ref{fig:main_norm_dimensions}, we can see that different models have a clear lead in different capabilities. Some examples include TerraMind-V1-Large for its multi-spectral and Multi-Temporal capabilities, or DINOv3-ViT-l-SAT for the $<10$ m GSD and RGB/NIR options. Although not ideal, given the goal of GeoFMs to perform well across all possible tasks, this is not surprising. In fact, based on their pretraining task, pretraining data distribution, target data, architecture, and size, some models will be more suitable to tackle specific use cases.


\subsection{Impact of Architecture, Size, and Pretraining Dataset}
Figure \ref{fig:bubble_plot} shows how architecture, model size, and pretraining dataset size relate to Core capability performance. Larger models consistently outperform their smaller counterparts, with ResNet-50 models showing notably poor performance regardless of pretraining data. However, Clay-V1 ViT-B achieves top Core performance with only 86M parameters. This is likely due to its diverse pretraining on ~70M heterogeneous EO samples (1-30m GSD) and smaller patch size of 8, which captures finer details at 4$\times$ computational cost compared to standard ViT models with patch size 16. Notably, ConvNeXt architectures pretrained on natural images adapt effectively to EO tasks through full finetuning, independent of model size or pretraining domain.

\subsection{Importance of Multi-Spectral Bands}
DINOv3-ViT-L-SAT and the ConvNeXt models are top performers in the \textbf{RGB/NIR} or the below 10 m GSD capability, which mostly includes high-resolution RGB data to perform the task. Looking at Figure \ref{fig:main_norm_dimensions}, we can see that despite TerraMind-V1-Large, Prithvi-EO-2.0-600-TL, and Clay-V1 ViT-B taking the podium in the Multi-Spectral-Dependent capability, the ConvNeXt models are still close in normalized performance. This is somewhat surprising because this capability only includes datasets that reported a statistically significant drop in performance once multi-spectral bands were removed to complete the task (see Appendix \ref{app:rgb_results}). However, when we look more in detail at the individual datasets' raw results in Figure \ref{fig:raw_results_per_dataset}, we see interesting insights emerging from datasets like NASA Burn Scars and PASTIS crop classification. Particularly, here there is a more marked difference (e.g. up to 10\%) between the models using all multi-spectral bands available and the ones using only RGB data. This confirms the importance of EO-specific models for particular applications in which going beyond RGB is the only way to achieve maximum performance.


\subsection{Model Performance Across Task Types}
Looking at the different task-focused capabilities, some architectural choices seem to bring clear advantages. For example, Clay-V1-ViT-B tops the Pixel-wise capability, which is consistent with its architecture. As already mentioned, it is the only GeoFM that adopts a smaller patch size, granting it an inherent advantage in retaining fine-grained spatial details crucial for these tasks. In the classification and detection capabilities, the DINOv3 and Imagenet ConvNeXt models achieve the highest rankings. In the detection tasks, however, most GeoFMs based on a Vision Transformer architecture show comparatively lower performance. This may be related to the choice of frameworks used for this task (i.e., Faster R-CNN and Mask R-CNN), which are known to favor models with convolutional inductive biases.

\subsection{Ablations}

Beyond the RGB versus Multi-Spectral ablation, we ran additional studies to assess the impact of some of our choices. We summarize these by reporting how rankings change, measured via Kendall tau distance—the percentage of model pairs with altered order (Figure~\ref{fig:rank_analysis_ablation}). Full details on all ablations are in Appendices~\ref{app:frozen_results}–\ref{app:linear_warm_up_results}.

\mypar{Frozen versus Fine-tuned Backbones.}
We first investigated whether freezing the backbone during training would cause performance degradation and ranking changes relative to full fine-tuning. Contrary to findings in~\cite{pangaea}, we observed that full fine-tuning consistently outperformed the frozen backbone approach, though the magnitude of performance drop varied across models. This resulted in ranking changes for over 20\% of model pairs, as shown in Figure~\ref{fig:rank_analysis_ablation}.

\mypar{Decoder Complexity.}
Next, we examined whether a simpler linear decoder could match the performance of a more complex UNet architecture while offering computational advantages. As detailed in Appendix~\ref{app:linear_results}, the linear decoder consistently underperformed across most datasets, leading to notable ranking changes.

\mypar{Multi-Temporal Processing.}
For datasets in the Multi-Temporal capability, we confirmed that leveraging multiple timestamps is crucial. Using only a single timestamp resulted in performance drops (Appendix~\ref{app:multitemp_results}) and produced the largest ranking perturbations among all ablations (Figure~\ref{fig:rank_analysis_ablation}).


\begin{figure}
\centering
\includegraphics[width=1.0\columnwidth]{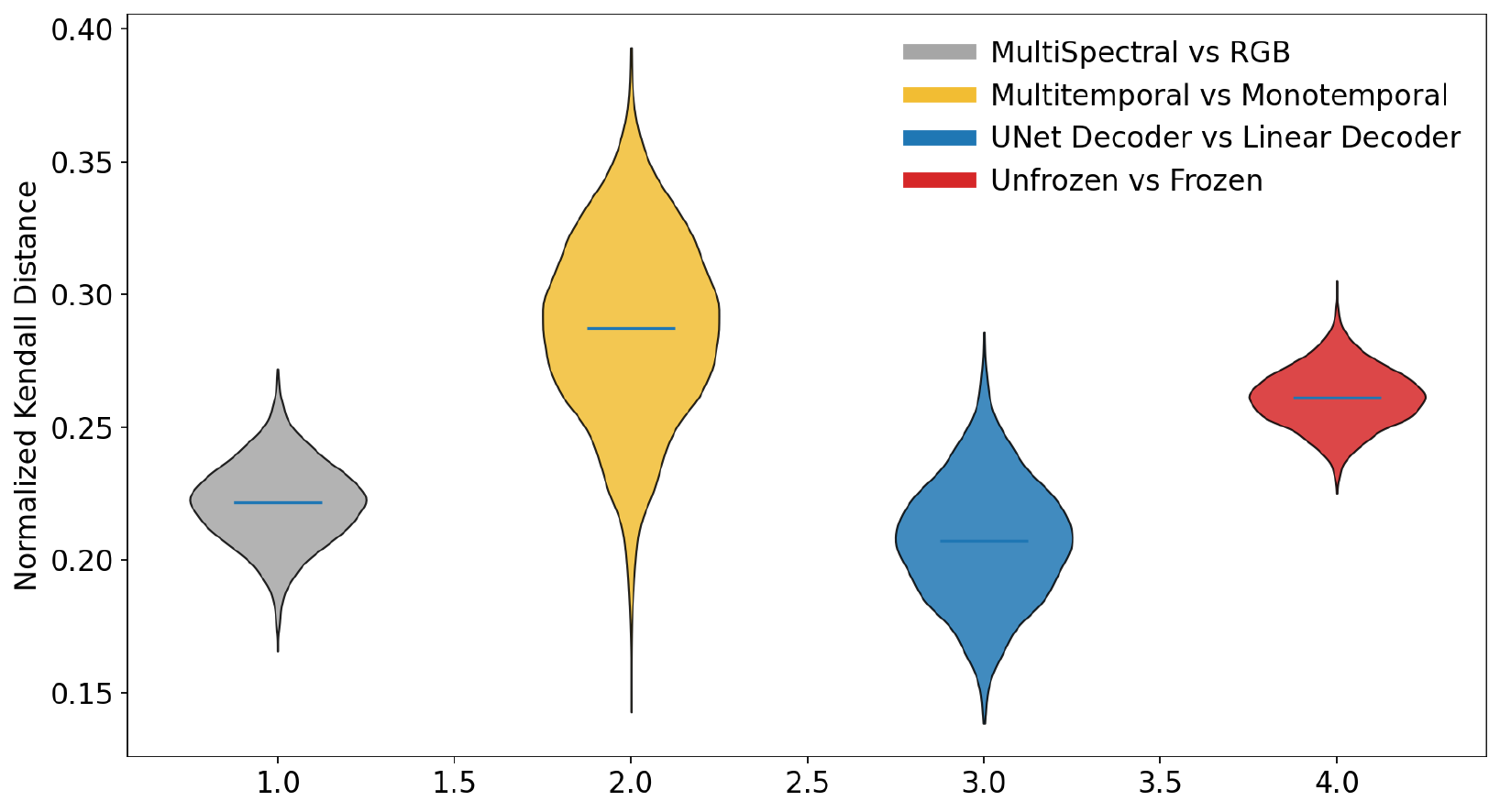}
\caption{Normalized Kendall tau distance showing the fraction of pairs where ranking is changing between 2 methods. Uncertainty was calculated through bootstrapping 2,000 times.}
\label{fig:rank_analysis_ablation}
\end{figure}

\section{Discussion and Limitations}

Foundation models for EO are advancing rapidly \cite{zhu2024foundations}, with a strong appeal for domain-specific applications. However, the field has yet to experience a paradigm shift comparable to that of generative modeling in text or computer vision. While our analysis confirms that larger models tend to perform better, EO has not exhibited the scaling laws observed in LLMs or image generation \cite{hoffmann2022training, li2024scalability}.

The LLM community benefits from established benchmarks and centralized infrastructure for tracking progress \cite{white2024livebench}. With GEO-Bench-2, we provide such a framework for the EO domain. A key advantage of the GEO-Bench-2 leaderboard is enabling users to rank models across various capabilities and settings, allowing the community to identify models suited to specific use cases and resource constraints. Our prescriptive protocol is centered on computational budget, repeated experiments, and generalizability, providing consistency while allowing flexibility in how the community adapts foundation models for different capabilities.

Recent work has shown that models pretrained on natural images remain competitive on EO tasks \cite{corley2024revisiting}, and well-tuned supervised models can match specialized FMs \citep{xuspecialized}. PANGAEA similarly found that most GeoFMs struggled to beat simpler models under frozen encoder settings \cite{pangaea}, with TerraMind being the first exception \cite{jakubik2025terramind}. 

Our experiments tell a more nuanced story. Under full fine-tuning settings, GeoFMs consistently outperformed simpler models (e.g., ResNet-50) but not larger models trained on natural images like ConvNeXt, particularly for high-resolution RGB-only tasks. However, GeoFMs demonstrated clear advantages when multi-spectral information is critical. This is precisely where GEO-Bench-2 can guide the field forward. By systematically evaluating models across diverse spectral bands, spatial resolutions, and temporal scales, we can identify where EO-specific architectures and pretraining truly matter. Our results align with DINOv3 findings \cite{dinov3}, where even the 7B RGB-only model could not surpass the 10$\times$ smaller Prithvi-EO-2.0 in crop classification. This highlights the value of multi-spectral capabilities that GEO-Bench-2 explicitly tests.

Rather than discouraging, these findings reveal opportunities. GEO-Bench-2 pinpoints where current GeoFMs underperform and provides a testbed for developing models that leverage EO's unique characteristics: multi-spectral data, temporal dynamics, and global coverage. These capabilities extend beyond what natural image pretraining can offer. This can guide the community toward further developing pretraining strategies optimized for multi-spectral and temporal data. For instance, our analysis revealed competitive results from ConvNeXt architecture, rarely used in EO, suggesting unexplored architectural directions. By systematically testing models across GEO-Bench-2's diverse tasks, the community can discover which innovations translate to real gains in the tasks that matter most: disaster response, environmental monitoring, and agriculture at scale.

\subsection{Limitations}

As with any benchmark, some arbitrary choices were necessary, but we believe they are reasonable and do not diminish the framework’s value. Our benchmark includes many leading models, though others remain untested, and some baseline decisions may not suit all model–dataset pairs. We encourage community contributions via the leaderboard. A SAR-specific capability is missing despite five SAR-inclusive datasets—future work should address this gap. Following \cite{rolf2021generalizable}, benchmarks should evolve toward real-world conditions with task-specific metrics and deployment-aware data, requiring cross-sector collaboration. Despite efforts toward global diversity, dataset coverage remains biased toward Europe and North America. Finally, predictive uncertainty is not assessed, though it is critical for real-world applications and should be prioritized.

{
    \small
    \bibliographystyle{ieeenat_fullname}
    \bibliography{main}
}

\newpage

\section{Appendix}

\subsection{Datasets}
\label{app:datasets}

The following sections introduce the different datasets as well as the respective preprocessing schemes chosen to generate the benchmark version of the dataset.

\subsubsection{BigEarthNet V2}
BigEarthNet V2 \cite{clasen2024reben} is an updated version of the popular BigEarthNet \cite{sumbul2019bigearthnet} multi-label classification dataset with Sentinel 1 and 2 imagery over Europe. The classification labels are derived from the CORINE Land Cover (CLC) labels. In BigEarthNet V2, the Sentinel 2 images were updated through a newer atmospheric correction scheme, the labels were revised with the most recent CLC labels, and a new geographic train, validation, and test split was introduced that reduces their spatial correlation. 

\begin{figure}[b]
    \centering
    \includegraphics[width=0.8\linewidth]{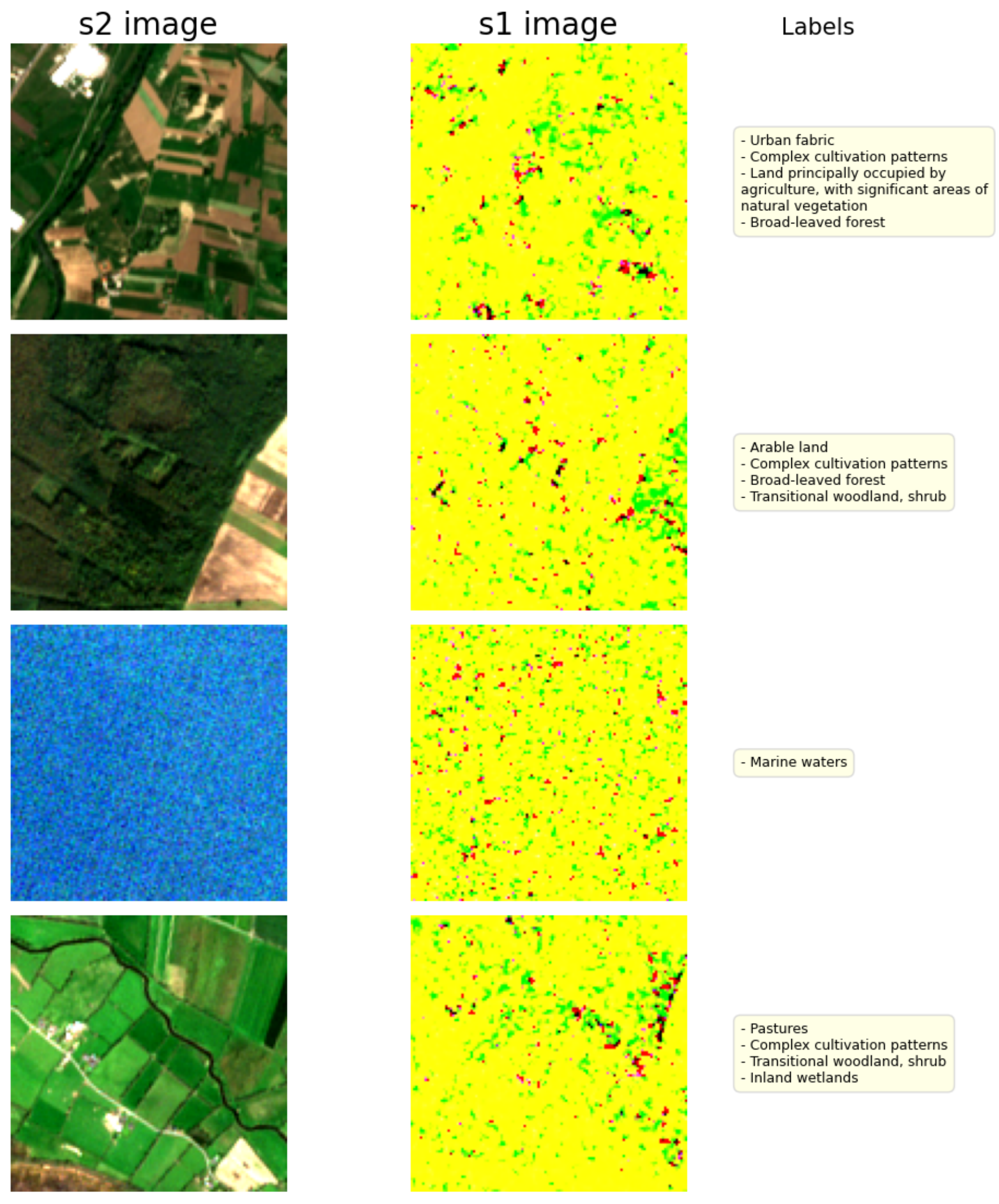}
    \caption{Training Set Examples for BigEarthNet V2 dataset.}
    \label{fig:benv2}
\end{figure}

\subsubsection{BioMassters}
The BioMassters dataset \cite{nascetti2023biomassters} consists of multi-temporal Sentinel 1 and 2 imagery over Finland and has pixel-wise above-ground biomass (AGB) annotations that were derived from airborne LiDAR measurements. The dataset was released with a designated train, validation, and test set but without geospatial information.

\begin{figure}[t]
    \centering
    \includegraphics[width=0.8\linewidth]{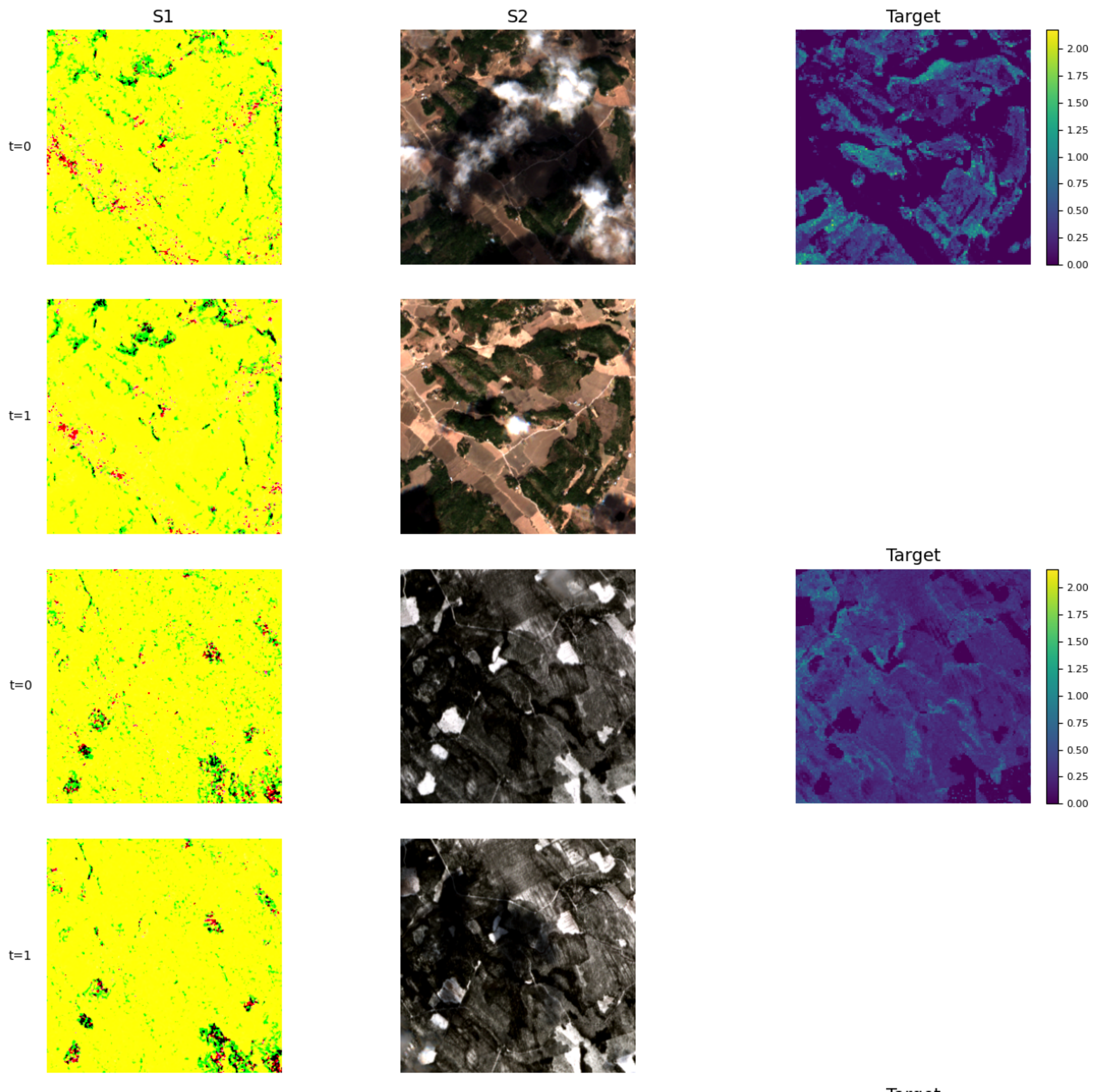}
    \caption{Training Set Examples for Biomassters dataset.}
    \label{fig:biomassters}
\end{figure}

\subsubsection{CaFFe}
The Calving Fronts and Where to Find them (CAFFe) dataset \cite{gourmelon2022calving} includes SAR imagery of glaciers from Antarctica, Greenland, and Alaska, with zone segmentation labels of rock, glacier, and ocean or ice melange. In the designated train, validation, and test set, the test set locations are disjoint from the other sets. The dataset was released as single-channel PNG files, but was reprocessed as TIF files with merged location information from a metadata table. 

\begin{figure}[b]
    \centering
    \includegraphics[width=0.8\linewidth]{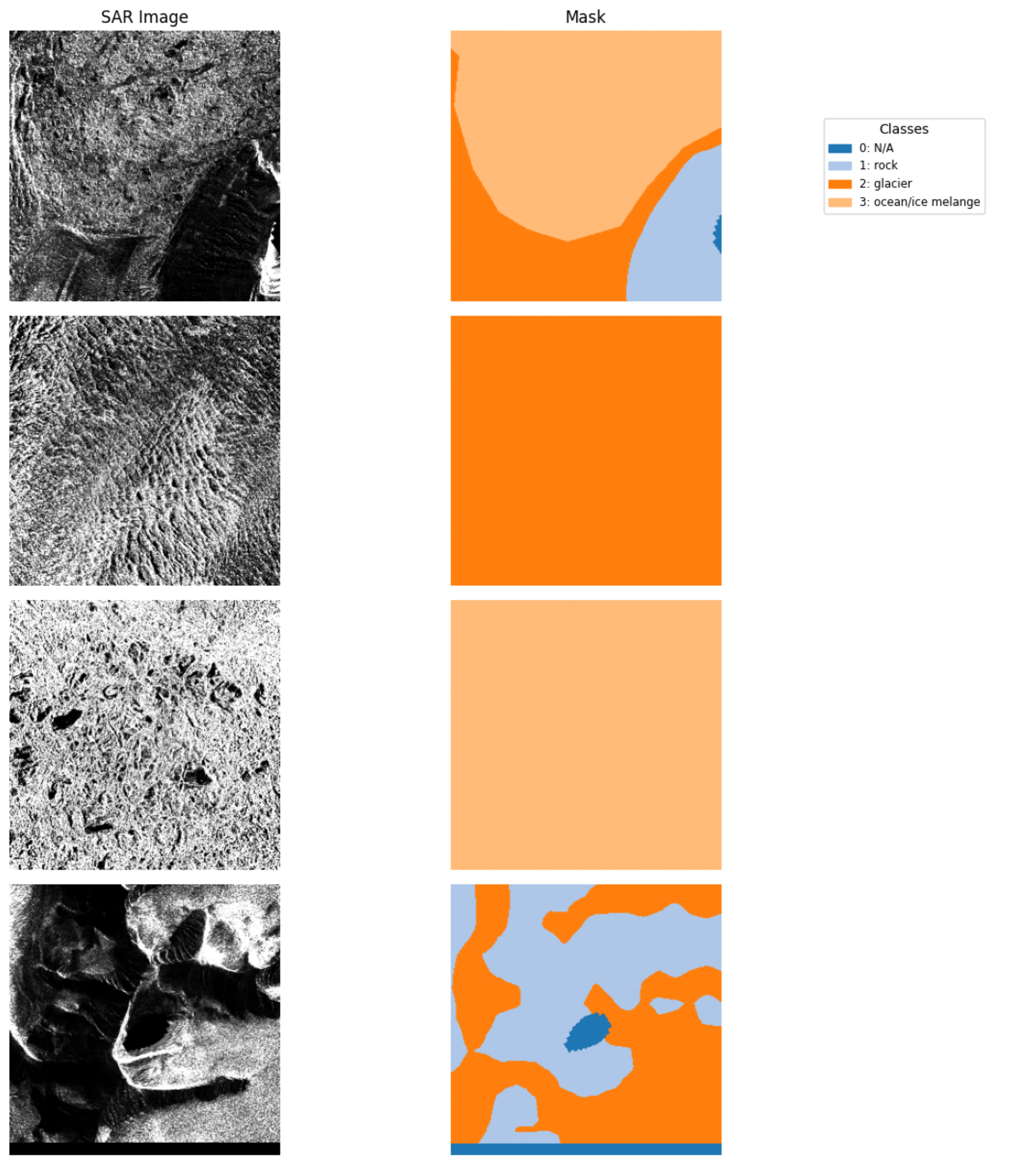}
    \caption{Training Set Examples for CaFFe dataset.}
    \label{fig:caffe}
\end{figure}

\subsubsection{Cloudsen12}
Introduced by \cite{cloudsen12+}, the Cloudsen12+ dataset forms a comprehensive global collection of Sentinel-1 and Sentinel-2 imagery for cloud and shadow detection. Samples were assigned to a train, validation, or test set based on a spatially stratified block split strategy. For our benchmark, we only select the processed samples with 512x512 pixels and the "high-quality" label tag.

\begin{figure}[h]
    \centering
    \includegraphics[width=0.7\linewidth]{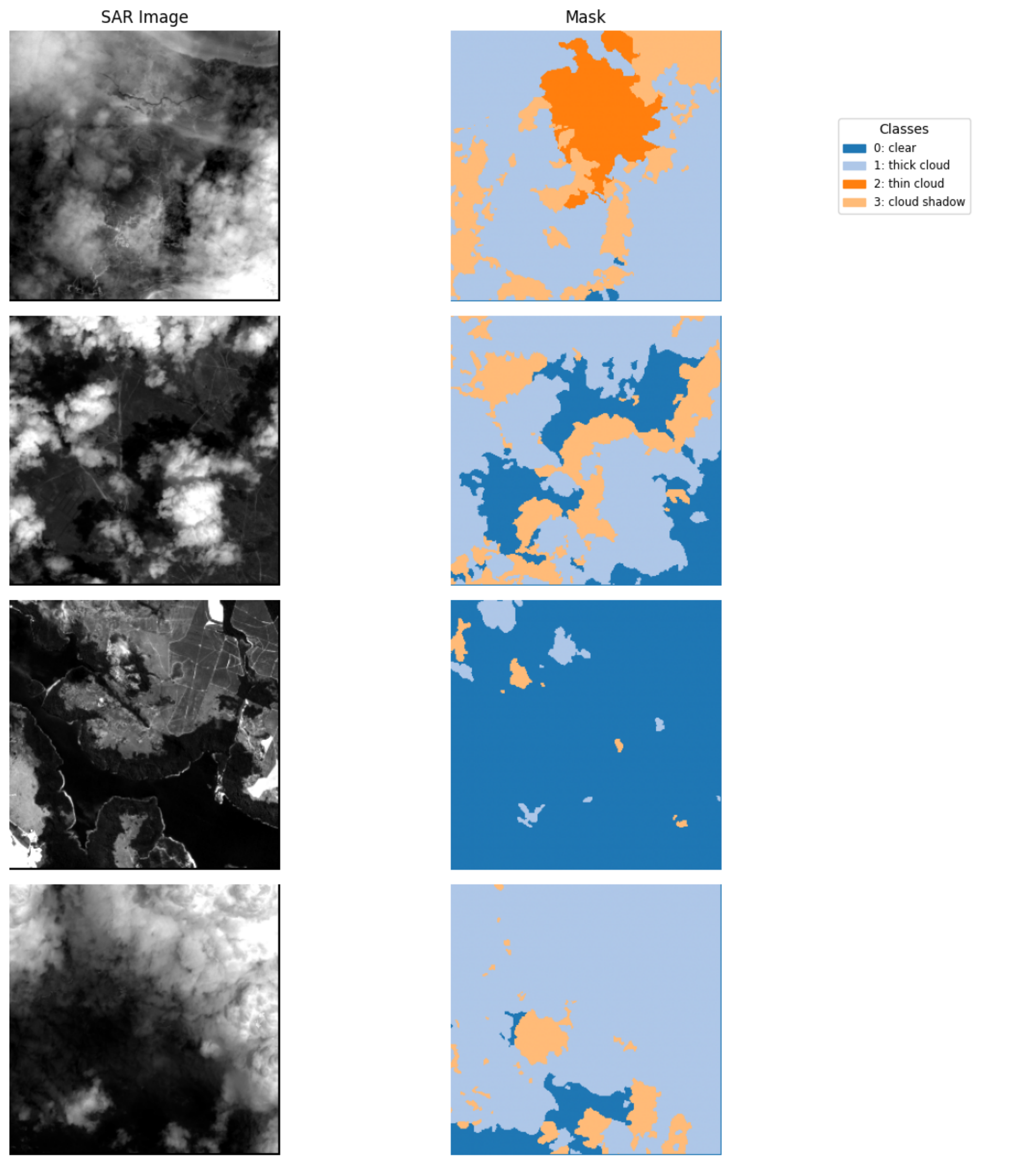}
    \caption{Training Set Examples for CloudSen12 dataset.}
    \label{fig:cloudsen12}
\end{figure}

\subsubsection{DynamicEarthNet}
The DynamicEarthNet dataset \cite{toker2022dynamicearthnet} is a global multi-temporal dataset of high-resolution Planet Lab imagery and Sentinel-2 imagery with seven land use and land cover (LULC) classes. 
\begin{figure}[b]
    \centering
    \includegraphics[width=0.8\linewidth]{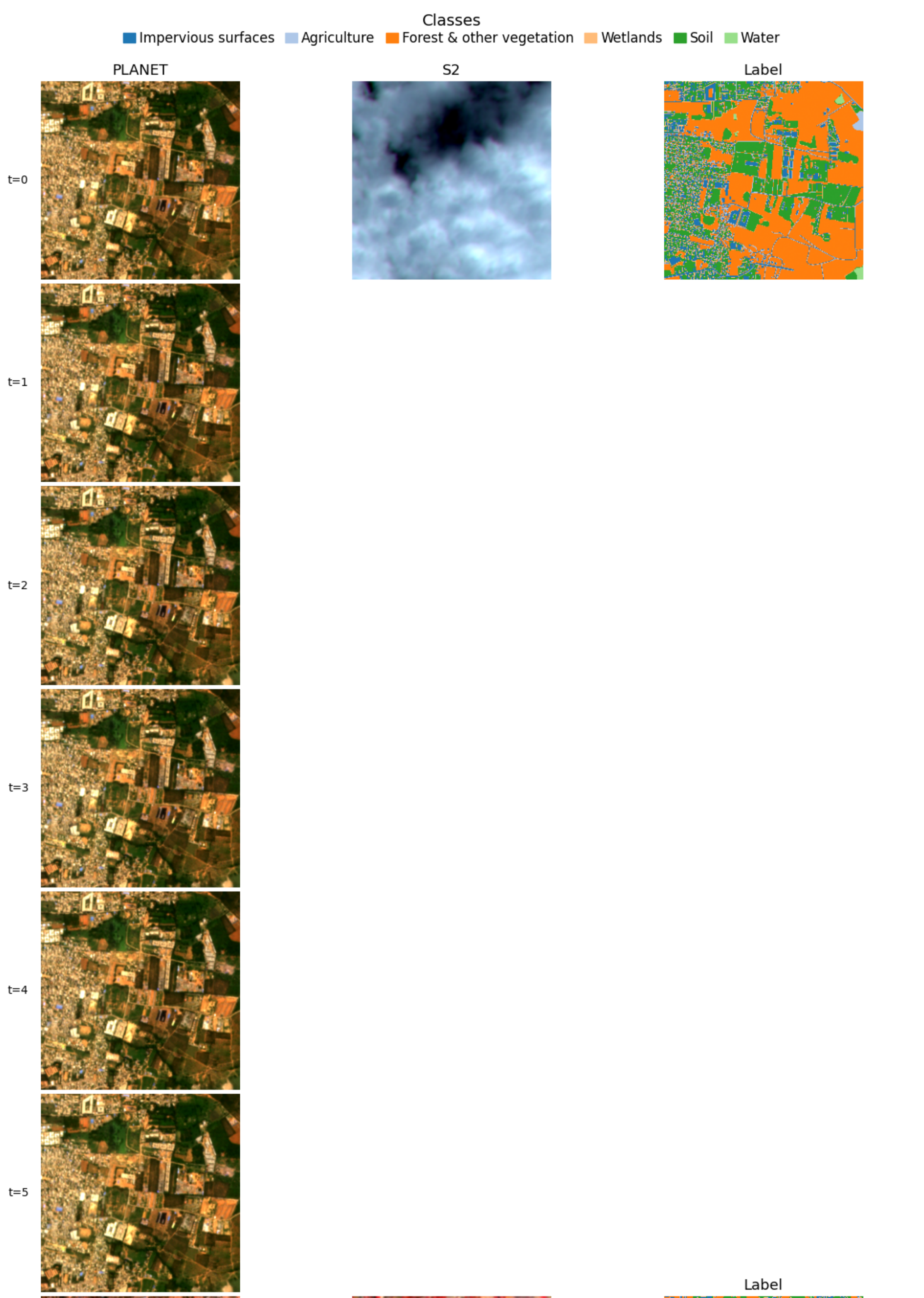}
    \caption{Training Set Examples for DynamicEarthNet dataset.}
    \label{fig:dynamic_earthnet}
\end{figure}
Based on 75 different areas of interest, across six continents, the dataset provides daily Planet imagery across these locations from 2018 to 2020. The samples released within the publicly available subset were processed in the following way. First, the 1024x1024 tiles were split into four 512x512 patches. Subsequently, an eight-by-eight grid binning strategy was used to assign samples to a train, validation or test set. The resulting splits neither overlap in space nor time.

\subsubsection{EverWatch}
The EverWatch \cite{garner2024everwatch} is a bird detection dataset with high-resolution imagery over the Everglades region. The aerial drone imagery was manually annotated with seven different bird species labels. The dataset comes with a designated train, validation, and test set and only partially complete geospatial information. The variable-sized tiles were resized to patches of size 512x512.

\begin{figure}[h]
    \centering
    \includegraphics[width=0.7\linewidth]{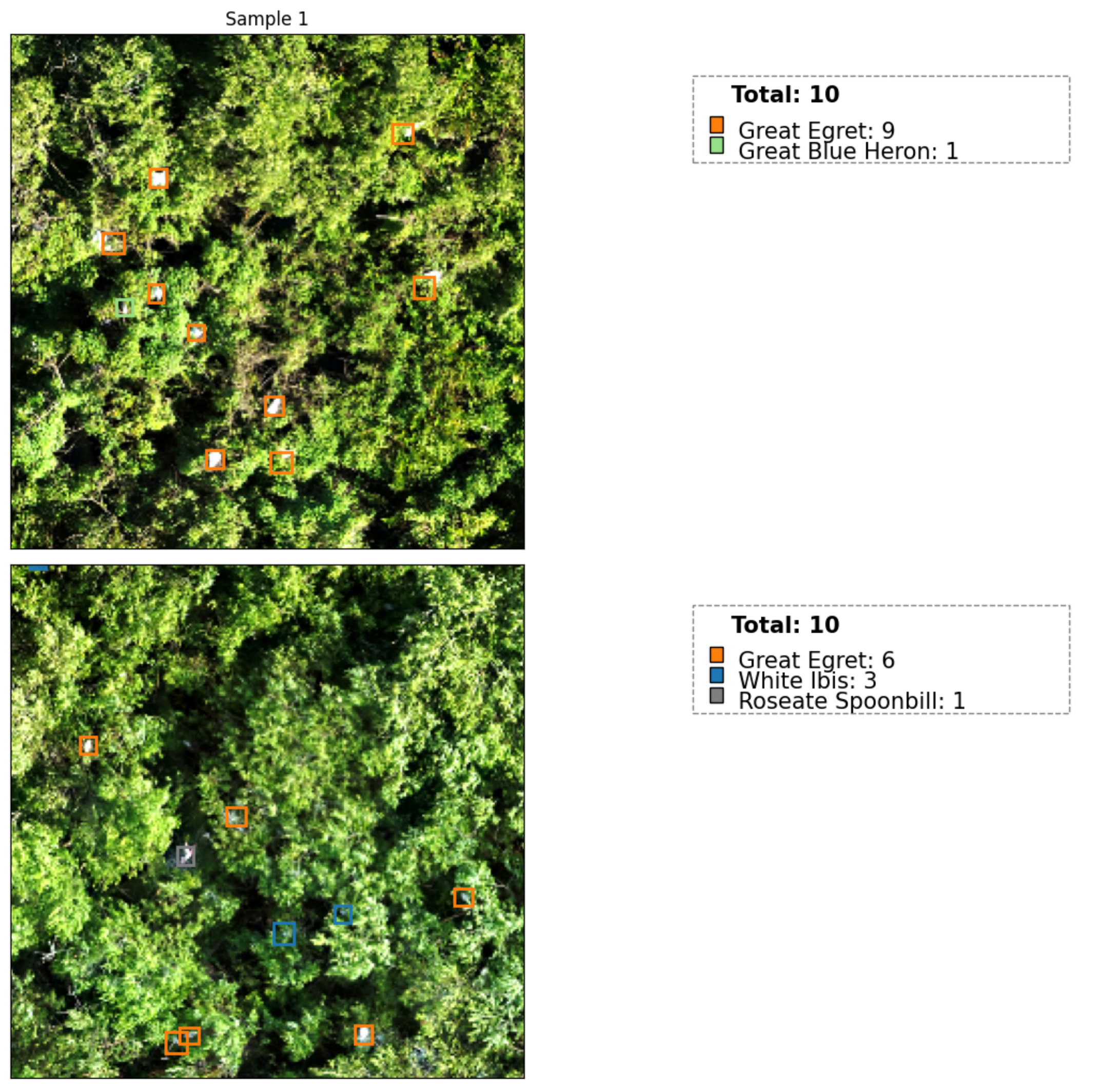}
    \caption{Training Set Examples for EverWatch dataset.}
    \label{fig:everwatch}
\end{figure}

\subsubsection{Fields of The World}
The Fields of the World (FTW) dataset \cite{kerner2025fields} is a global dataset of Sentinel-2 imagery for agricultural field boundary delineation. 
\begin{figure}[b]
    \centering
    \includegraphics[width=0.6\linewidth]{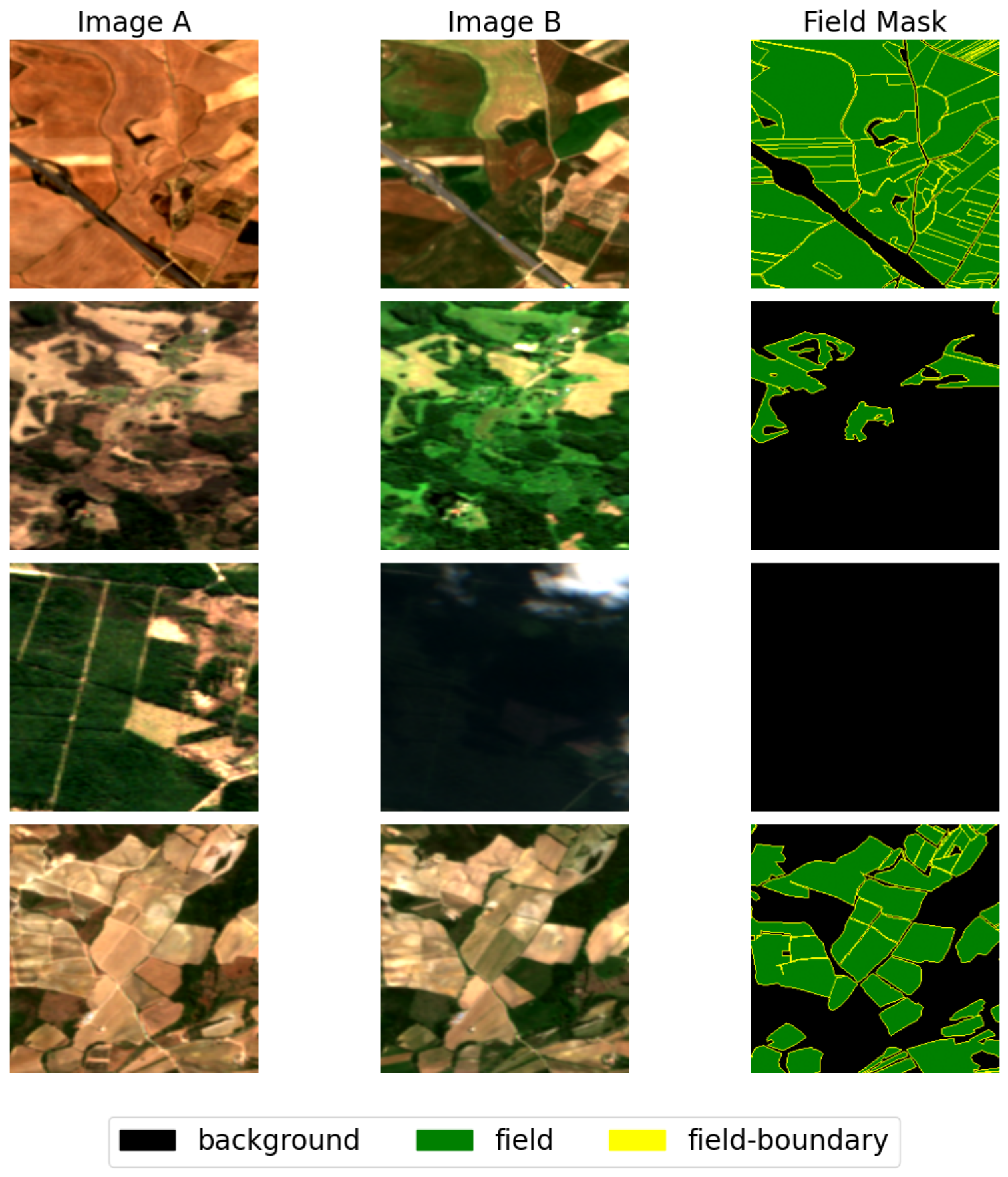}
    \caption{Training Set Examples for FTW dataset.}
    \label{fig:ftw}
\end{figure}
Each sample consists of two temporal views from different time stamps and a field mask. The dataset was filtered by country to only include data consistent with the open CC-BY-SA license. For this subset, the original train, validation, and test split was used.

\subsubsection{FLAIR2}
The French Land cover from Aerospace ImageRy (FLAIR) Version 2 dataset \cite{garioud2023flair} is an updated version of FLAIR \cite{garioud2022flair} that now also contains Sentinel-2 time-series data next to the high-resolution aerial imagery with thirteen semantic land cover classes across France. The new version also includes an updated test split from distinct spatial domains across the country.

\begin{figure}[t]
    \centering
    \includegraphics[width=0.65\linewidth]{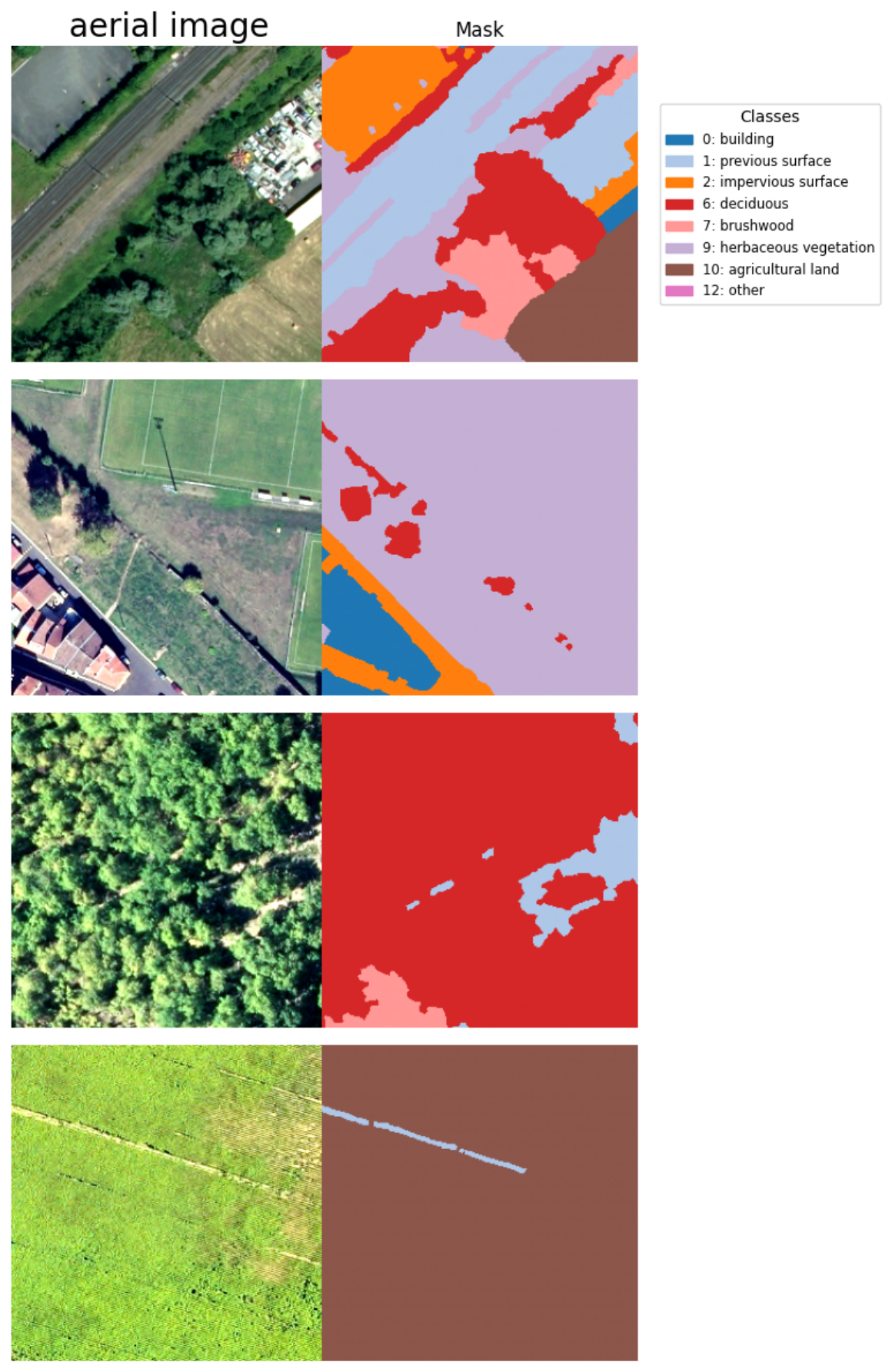}
    \caption{Training Set Examples for FLAIR2 dataset.}
    \label{fig:FLAIR2}
\end{figure}

\subsubsection{KuroSiwo}
The KuroSiwo dataset \cite{bountos2024kuro} is a global dataset of SAR imagery from a broad range of global flooding events with annotations for permanent water and flood water. We use the designated train, validation, and test set.

\begin{figure}[b]
    \centering
    \includegraphics[width=0.8\linewidth]{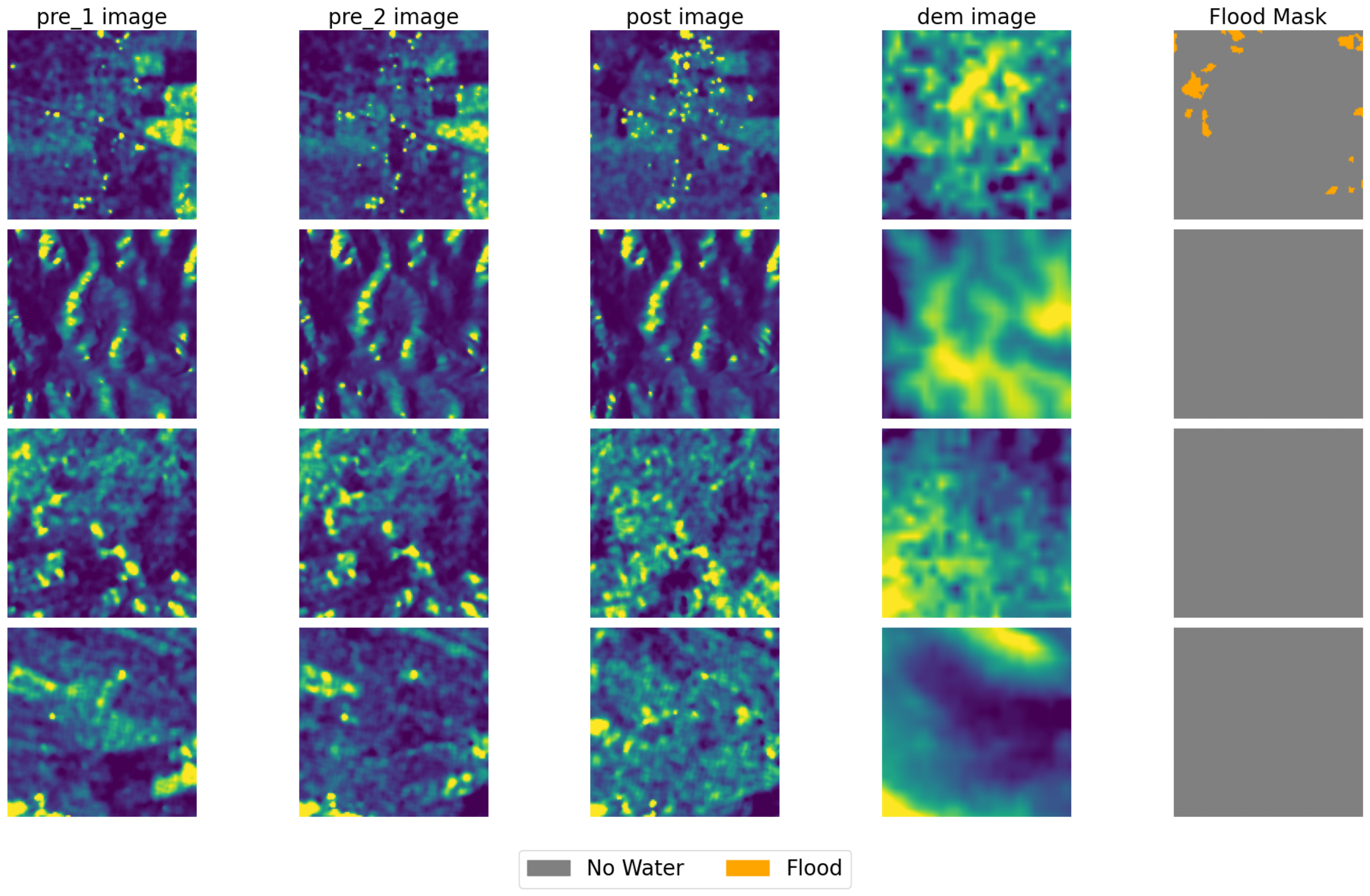}
    \caption{Training Set Examples for KuroSiwo dataset.}
    \label{fig:kuro_siwo}
\end{figure}

\subsubsection{PASTIS-R}
The Panoptic Agricultural Satellite Time Series (PASTIS) dataset \cite{garnot2022multi} is a multi-temporal dataset with Sentinel-1 SAR and Sentinel-2 imagery with crop type annotations across France. We utilize the PASTIS-R version, which is a superset of the original PASTIS dataset and includes the SAR imagery. The dataset comes with a designated train, validation, and test split.



\begin{figure}[t]
    \centering
    \begin{subfigure}{0.45\textwidth}
        \centering
        \includegraphics[width=0.8\linewidth]{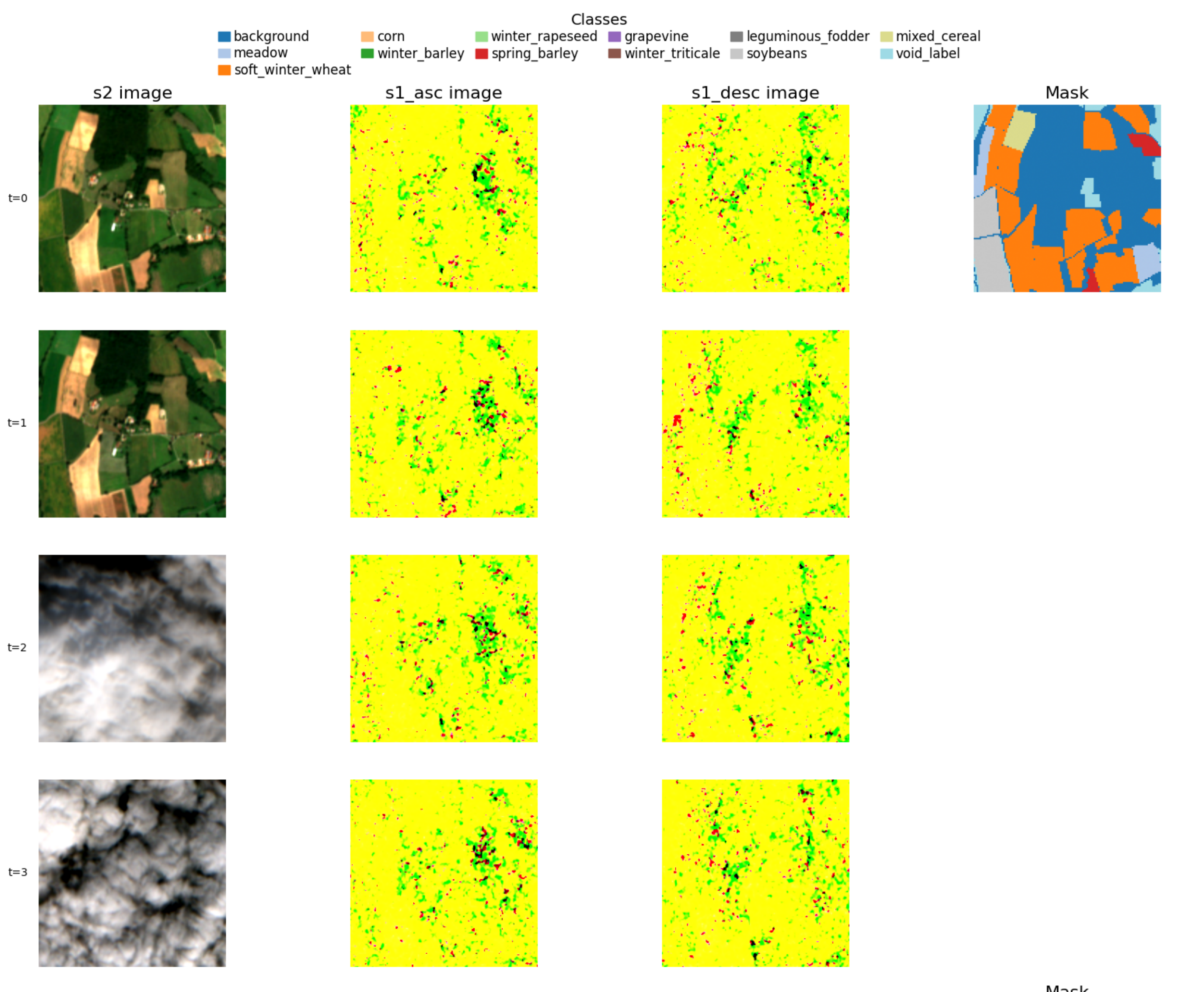}
        \caption{Training Set Examples for PASTIS Semantic Segmentation dataset.}
        \label{fig:pastis_segmentation}
    \end{subfigure} 
    \begin{subfigure}{0.45\textwidth}
        \centering
        \includegraphics[width=0.8\linewidth]{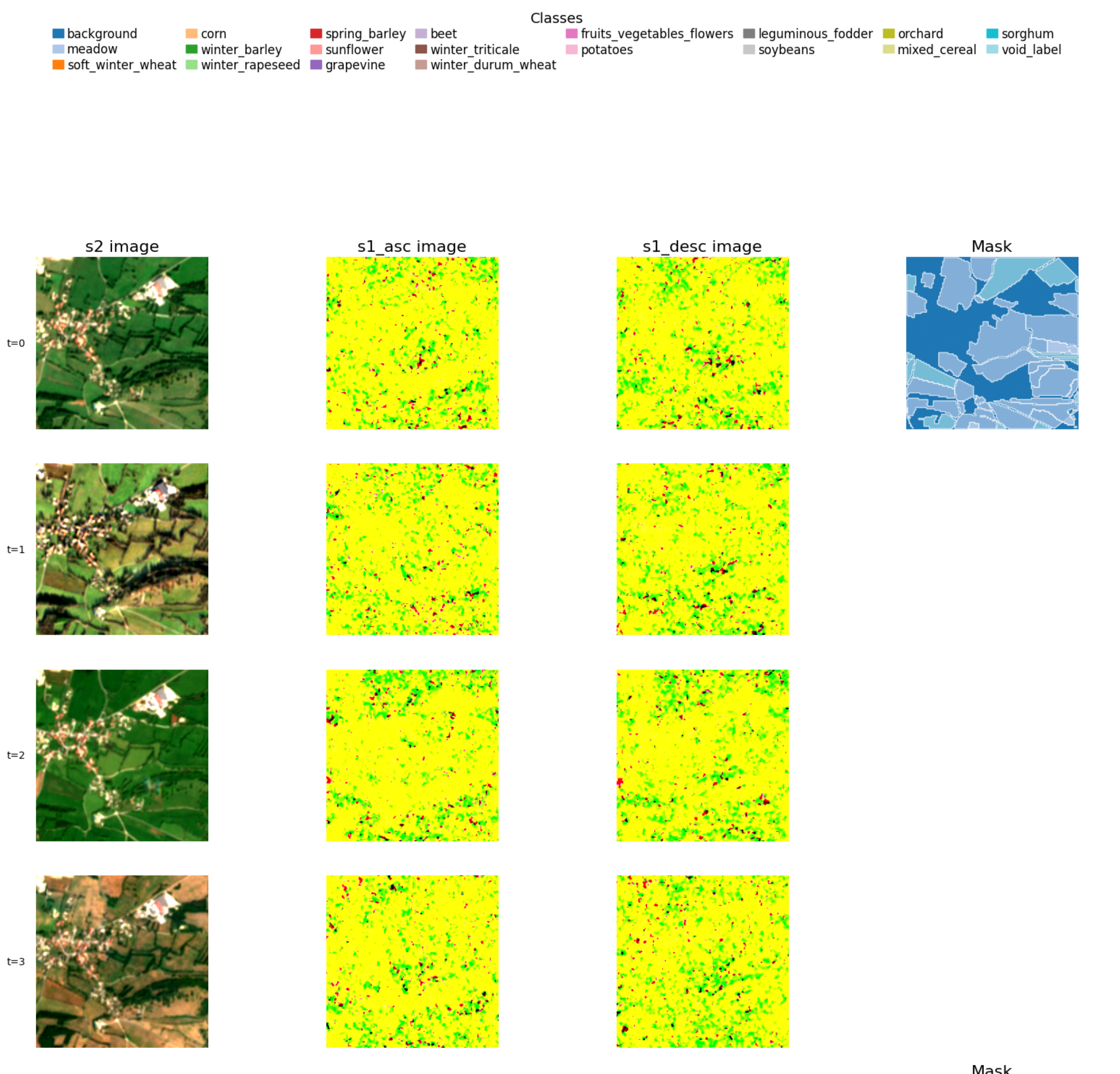}
        \caption{Training Set Examples for PASTIS Instance Segmentation dataset.}
        \label{fig:pastis_instance_segmentation}
    \end{subfigure} 
    \caption{Training Set Examples for PASTIS datasets.}
    \label{fig:pastis}
\end{figure}

\subsubsection{SpaceNet2}
The second edition of the SpaceNet dataset series \cite{van2021spacenet} consists of high-resolution aerial imagery across four different cities (Las Vegas, Paris, Shanghai, Khartoum) with binary building footprint annotations. As a competition dataset, the publicly available data that includes annotations is only designated for training. To have separate training and evaluation sets, we use the following procedure: For each city, we use a checkerboard style split, inspired by MOSAIKS \cite{rolf2021generalizable}, that overlays a grid structure and all samples within a grid are assigned to a train, validation and test set such that their percentage of the total samples is roughly 70-20-10 respectively.

\begin{figure}[t]
    \centering
    \includegraphics[width=0.7\linewidth]{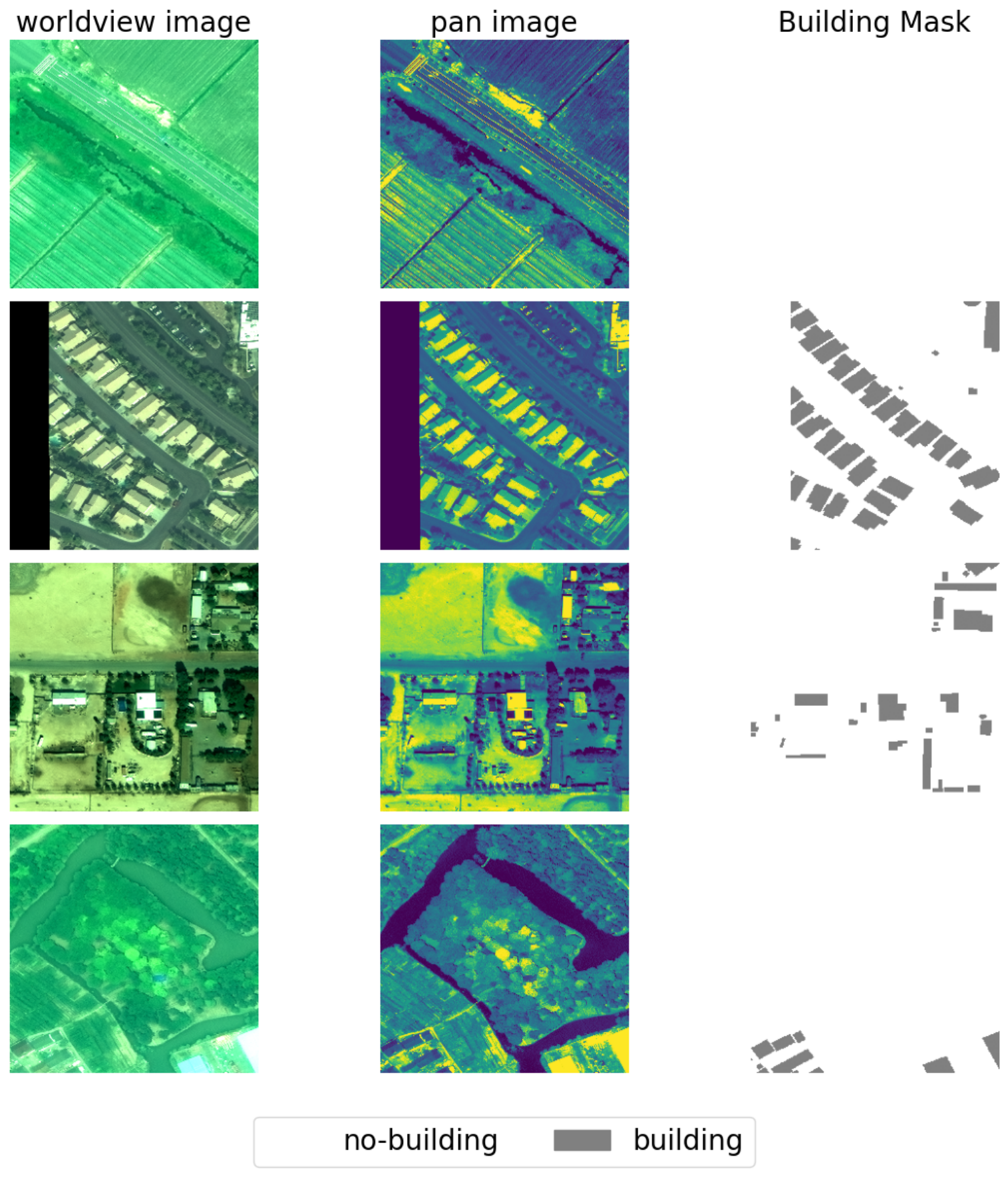}
    \caption{Training Set Examples for SpaceNet2 dataset.}
    \label{fig:spacenet2}
\end{figure}

\subsubsection{SpaceNet7}
The seventh edition of the SpaceNet dataset series \cite{van2021multi} is a multi-temporal building footprint dataset of 101 different locations across the globe. For each location, there is a sequence of Planet Lab Dove RGB imagery with fine-grained building annotations. While the intended purpose was to track building changes over time, we cast it as a static building segmentation dataset. The original 1024x1024 tiles were separated into 4 512x512 patches, and samples were assigned to a train, validation, or test split such that their locations are disjoint and their percentage of total samples is roughly 70-20-10 respectively.

\begin{figure}[b]
    \centering
    \includegraphics[width=0.5\linewidth]{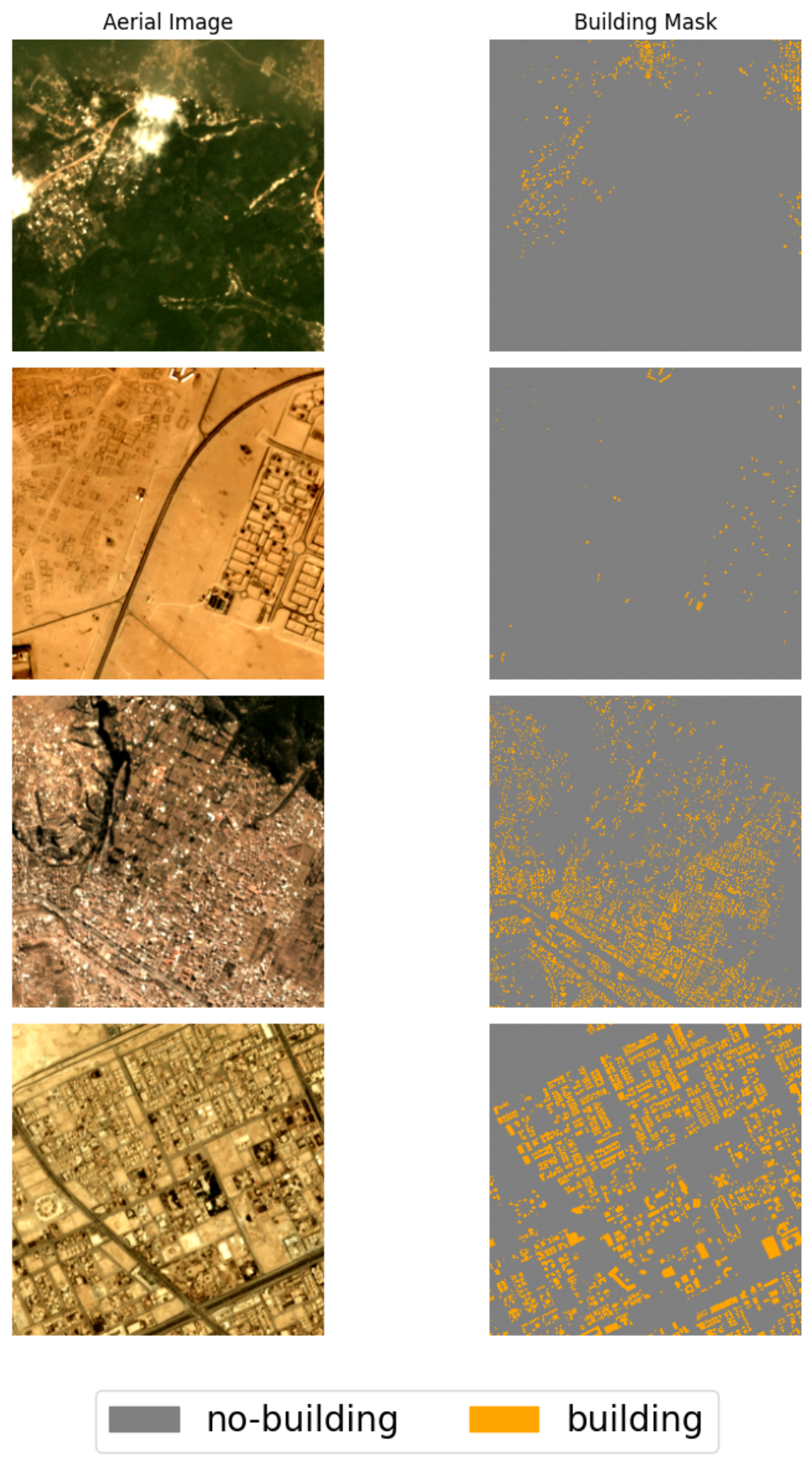}
    \caption{Training Set Examples for Spacenet7 dataset.}
    \label{fig:spacenet7}
\end{figure}

\subsubsection{TreeSatAI}
The TreeSatAI dataset \cite{ahlswede2022treesatai} originally included single-timestamp aerial, Sentinel-1, and Sentinel-2 imagery with corresponding multi-label tree species classes. \cite{astruc2024omnisat} extended the dataset to include a time series for the Sentinel modalities. 

\begin{figure}[t]
    \centering
    \includegraphics[width=0.4\linewidth]{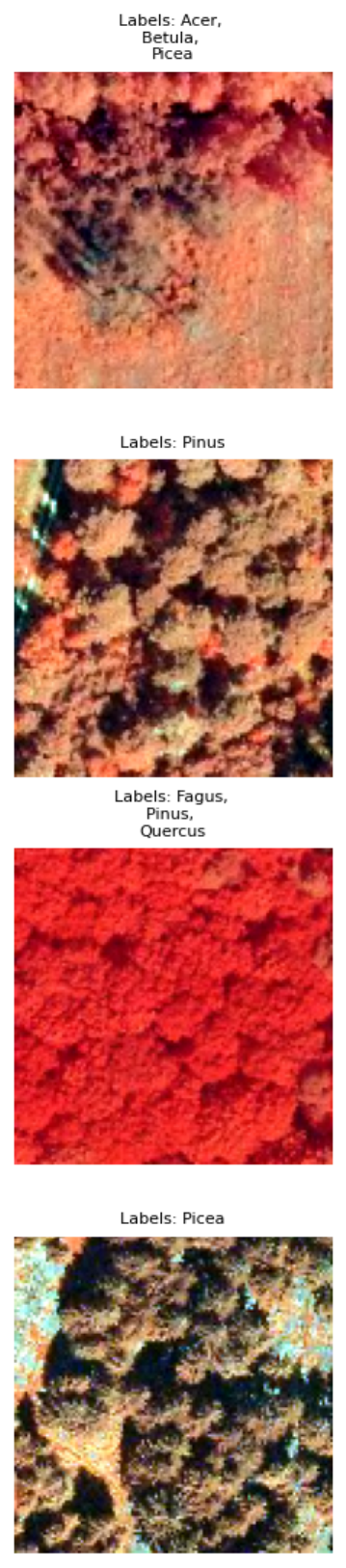}
    \caption{Training Set Examples for TreeSatAI dataset.}
    \label{fig:treesatai}
\end{figure}

The samples again were assigned to a train, validation, or test set with a checkerboard style split strategy, where a 10x10 grid was overlaid on the dataset area of northern Germany, and samples within each block belong to one split such that there is roughly a 70-20-10 distribution across splits.

\subsubsection{NZCattle}
The NZCattle is a dataset included in GEO-Bench \cite{geobenchv1}. It includes high-resolution aerial RGB images to detect cattle in New Zealand. Originally, the dataset was presented as a segmentation task, which here we repurpose it as an object detection task. Original splits were retained.

\begin{figure}[t]
    \centering
    \includegraphics[width=0.9\linewidth]{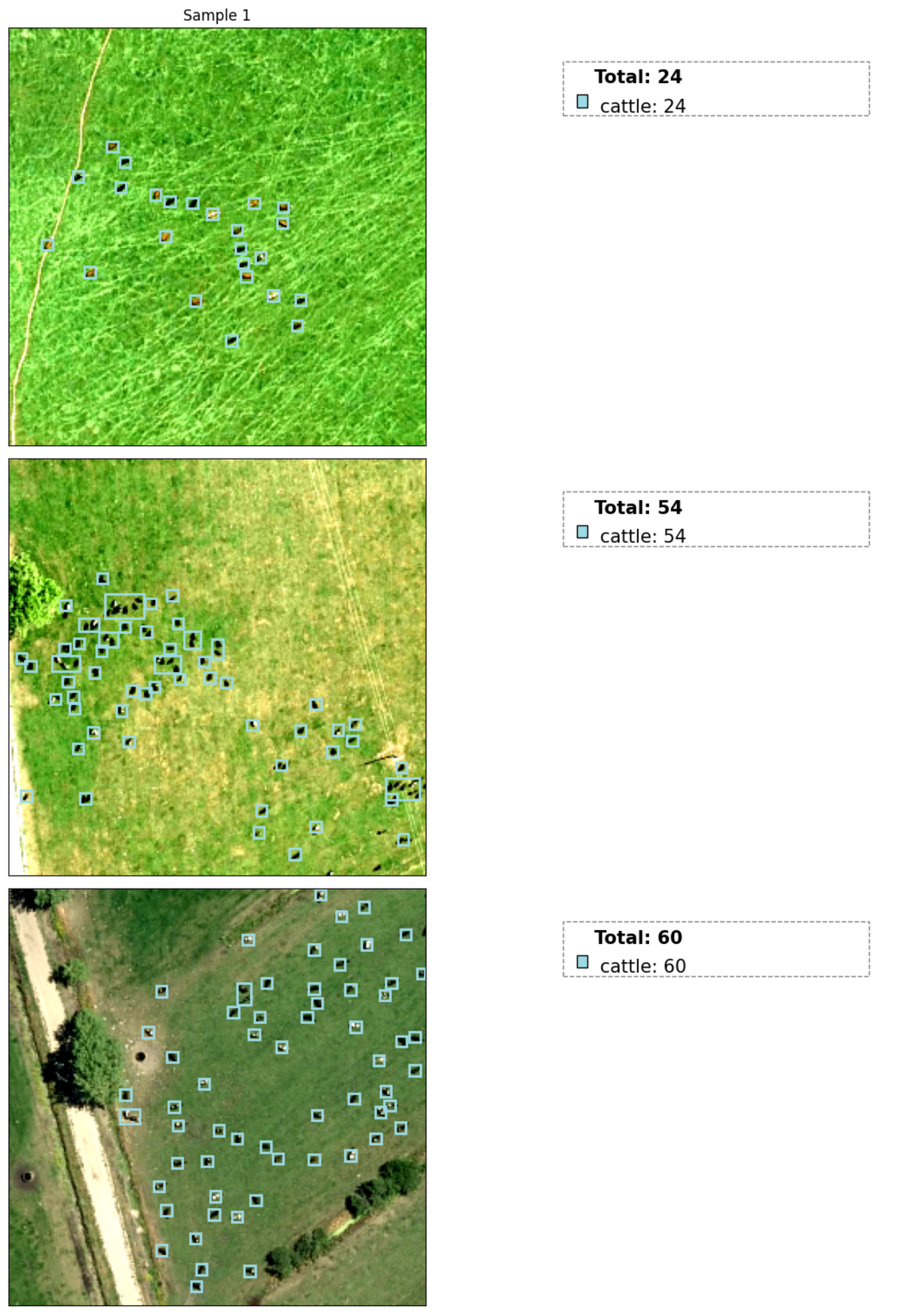}
    \caption{Training Set Examples for NZCattle dataset.}
    \label{fig:pastis_segmentation}
\end{figure}

\subsubsection{BurnScars}

This dataset contains Harmonized Landsat and Sentinel-2 imagery of burn scars and the associated masks for the years 2018-2021 over the contiguous United States \cite{prithvi2}. There are 804 512x512 images. For the benchmark we took the original dataset and just repurposed it in TACO format with no modifications.

\begin{figure}[h]
    \centering
    \includegraphics[width=0.9\linewidth]{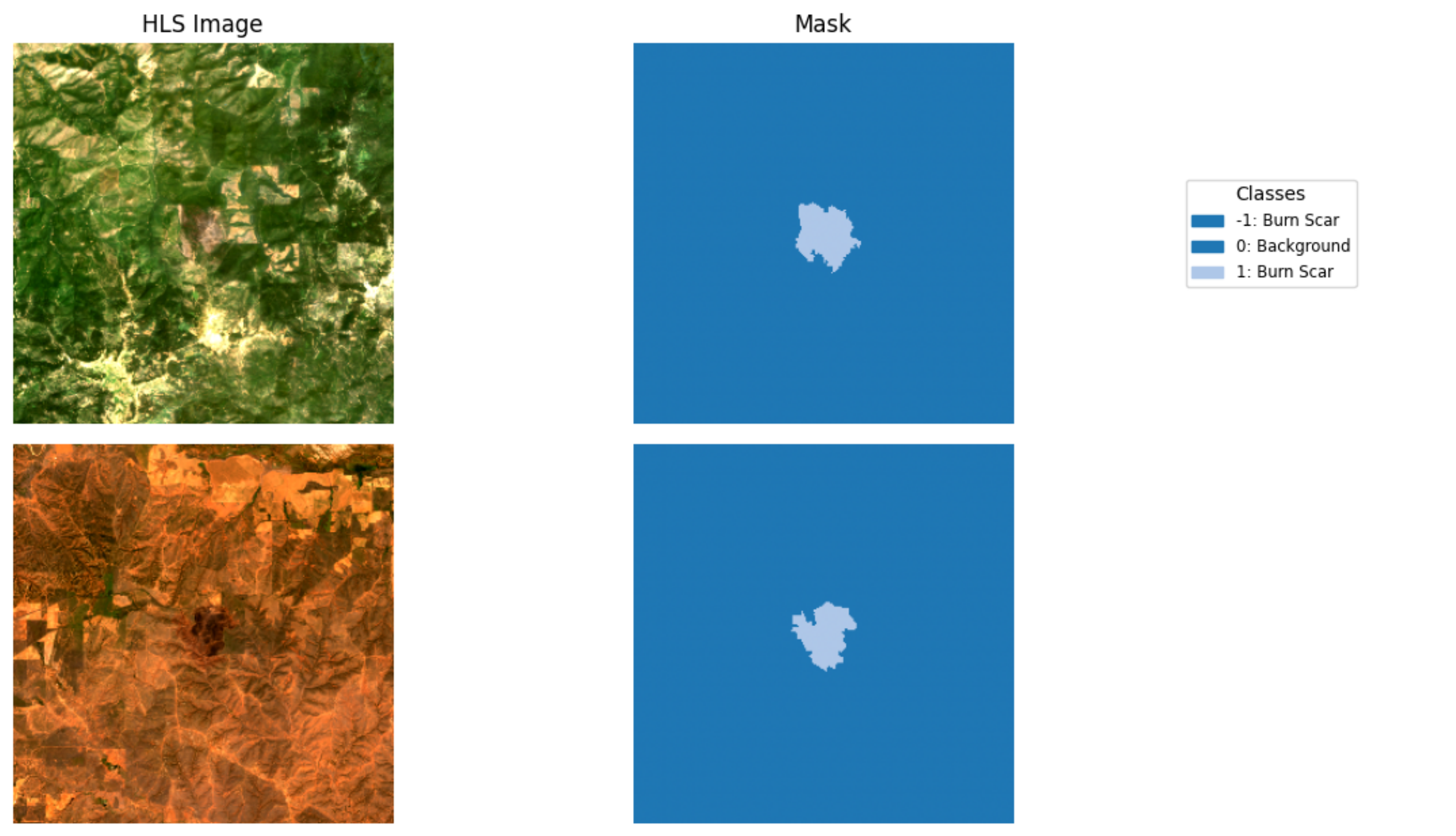}
    \caption{Training Set Examples for BurnScars dataset.}
    \label{fig:burn_scars}
\end{figure}

\subsubsection{Substation}
This dataset was curated by TransitionZero and sourced from publicly available data repositories, including OpenStreetMap and Copernicus Sentinel data \cite{substations}. The dataset consists of Sentinel-2 228x228 pixels images from 27k+ global locations; the original task was to semantically segment power substations. For the benchmark, we converted the original labels into instance segmentation labels. Furthermore, as most locations had 4-5 images taken at different timepoints (i.e., revisits), but beyond clouds occlusions, there is, in principle, no need for multiple timestamps to complete the task, we created a cloud-free composite for each location. Data was sub-sampled for each split to reduce computational requirements.

\begin{figure}[t]
    \centering
    \includegraphics[width=0.8\linewidth]{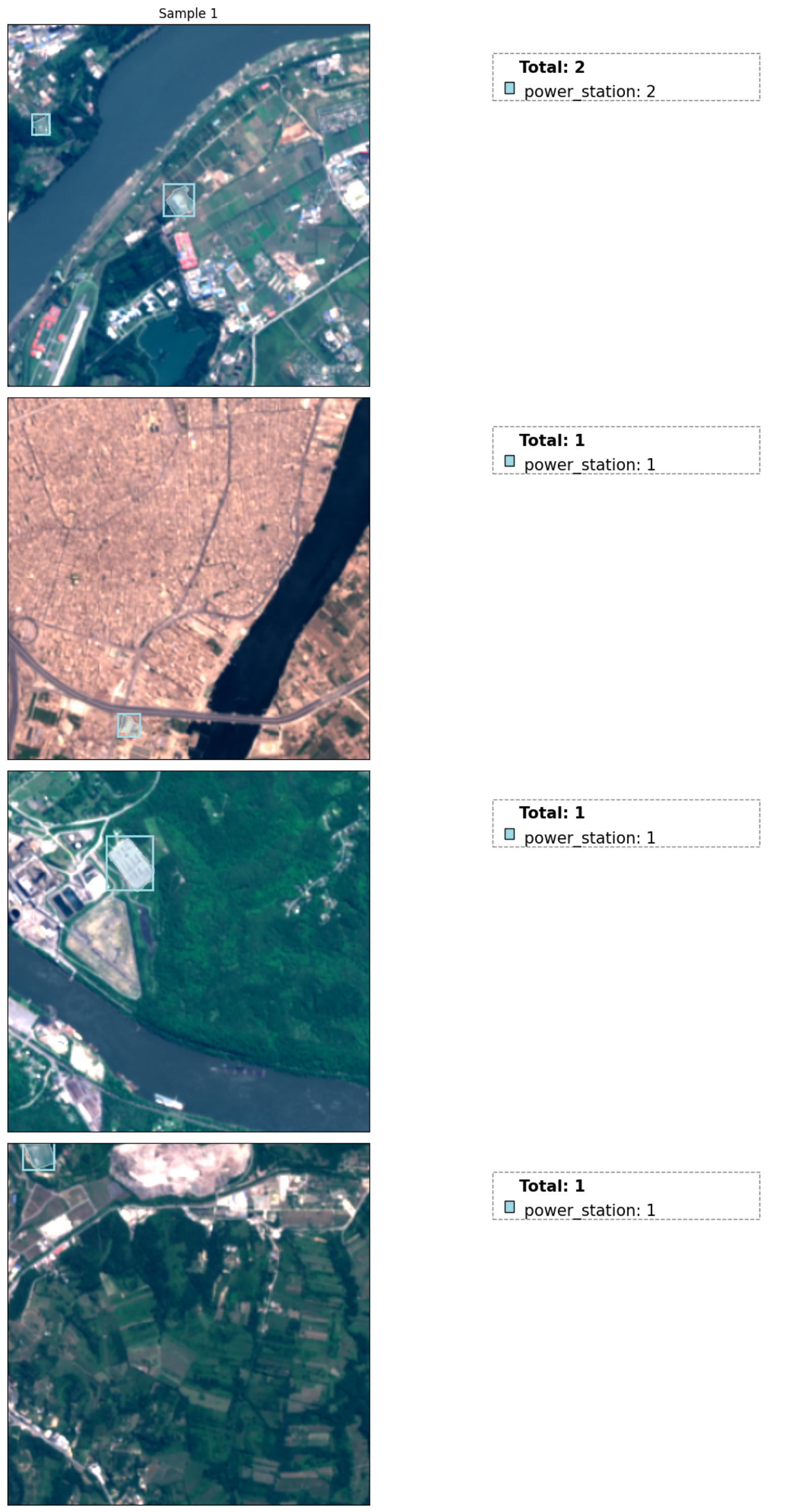}
    \caption{Training Set Examples for Substation dataset.}
    \label{fig:substation}
\end{figure}

\subsubsection{So2sat}
So2Sat is a dataset designed for multimodal classification. The aim is to classify local Climate Zones globally. We downloaded the modified version of this dataset from GEO-Bench \cite{geobenchv1} and repurposed it in TACO format with no modifications.

\begin{figure}[h]
    \centering
    \includegraphics[width=0.7\linewidth]{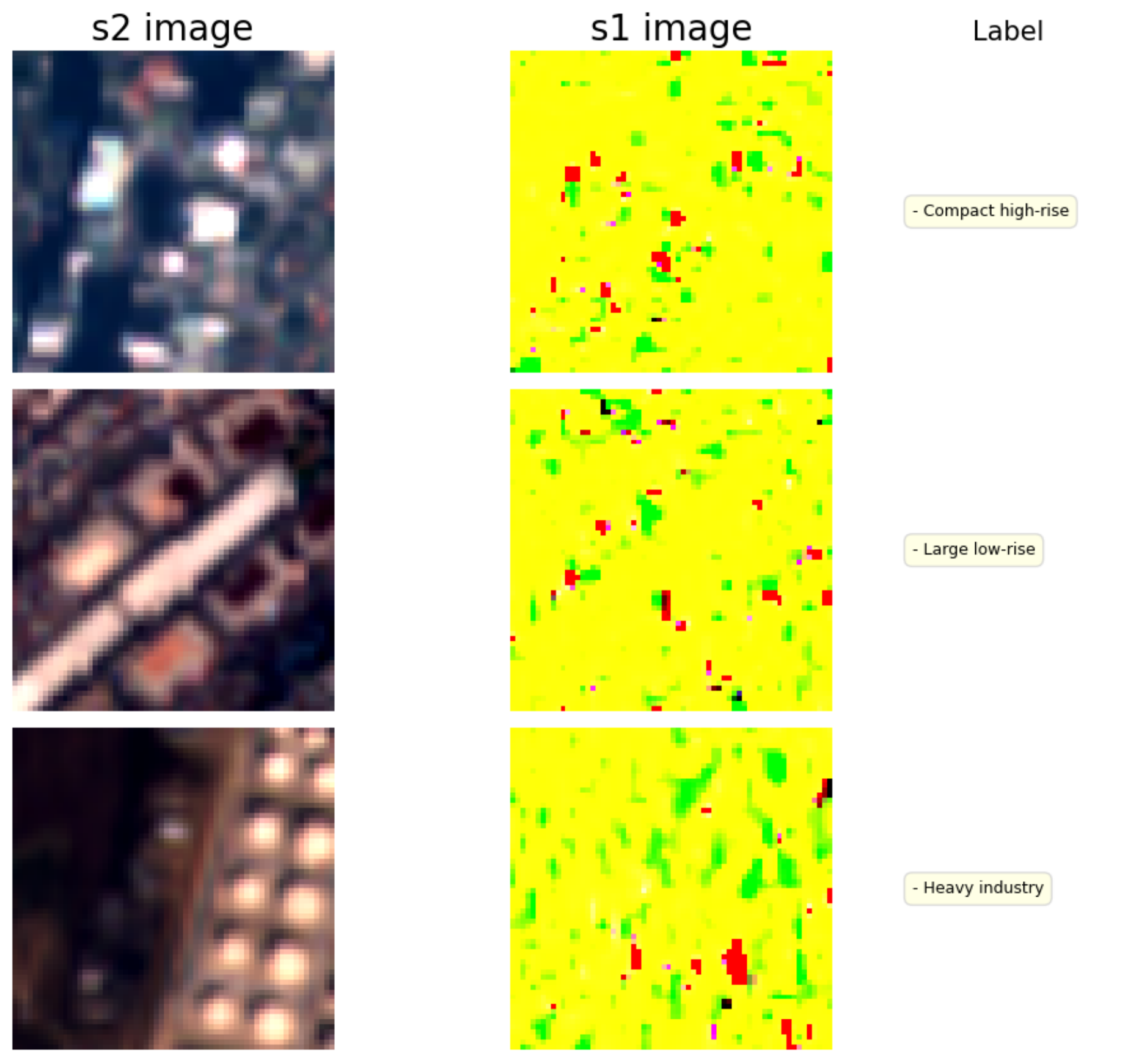}
    \caption{Training Set Examples for So2Sat dataset.}
    \label{fig:so2sat}
\end{figure}

\subsubsection{Forestnet}
Forestnet is a curated dataset of Landsat 8 satellite images of known forest loss events paired with driver annotations from expert interpreters in South Asia. We downloaded the modified version of this dataset from GEO-Bench  \cite{geobenchv1} and repurposed it in TACO format with no modifications.

\begin{figure}[t]
    \centering
    \includegraphics[width=0.55\linewidth]{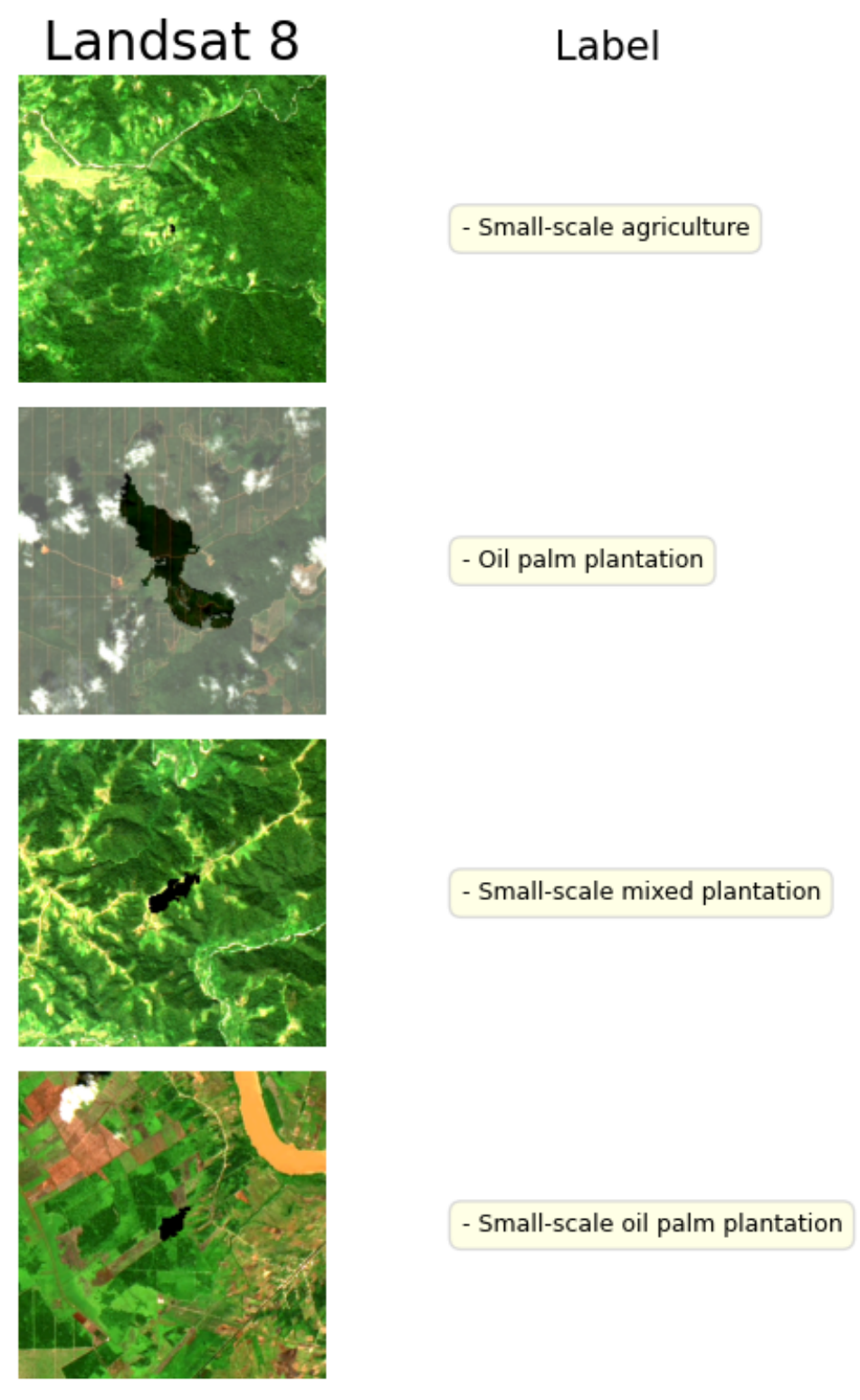}
    \caption{Training Set Examples for Forestnet dataset.}
    \label{fig:forstnet.}
\end{figure}

\subsubsection{Dataset Geographical Distribution by continent}

Figure \ref{fig:geospatial_distribution_continets} shows the geographical distribution of dataset samples across continents.

\setcounter{figure}{23}
\begin{figure}[htb]
    \centering
    \includegraphics[width=0.8\linewidth]{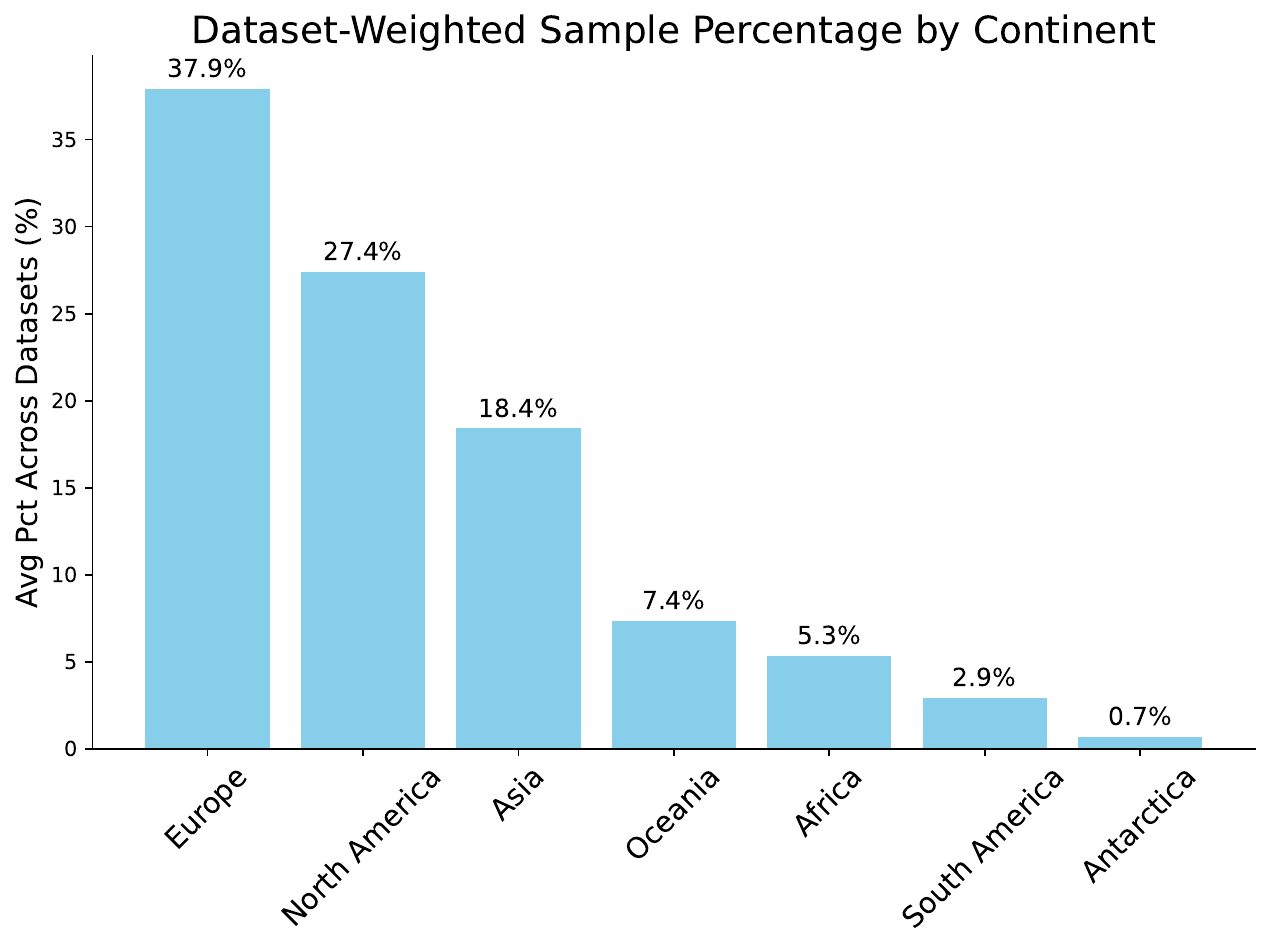}
    \caption{Distribution of samples weighted by source dataset size across continents. Each dataset's continent distribution is normalizing by the dataset size before aggregating across all dataset.}
    \label{fig:geospatial_distribution_continets}
\end{figure}

\subsubsection{Dataset Statistics}
Z-score normalization statistics were computed modality and band-wise for each dataset on the training set split and subsequently used to normalize all inputs. For the BioMassters dataset that contains continuous pixel-wise regression labels, z-score normalization was also used for the labels.
   
\subsection{Detailed experimental setup}
\label{app:experimental_setup}
To ensure reliable results, the evaluation protocol employs a combination of hyperparameter optimization (HPO) followed by repeated seed experiments.
For segmentation, regression, and classification tasks, we conduct 16 trials during the HPO phase. Due to higher computational requirements, detection experiments use 10 trials. After the HPO process for a given model and dataset combination, the set of parameters yielding the best validation set metric is selected and subsequently used for repeated experiments with five different random seeds. The final performance is then calculated on the test for each seed. Both HPO and repeated experiments are configured through the TerraTorch Iterate library, which automates the benchmarking workflow. HPO employs Optuna \cite{optuna} with Bayesian optimization and a Tree Structured Parzen Estimator (TPE) for efficient sampling. The general optimization search space was fixed across all models and datasets:
\begin{itemize}
 \item Learning Rate: A real value ranging from $1\text{e-}6$ to $1\text{e-}3$.
 \item Batch Size: An integer value selected from the list: $\{8, 16, 32\}$.
\end{itemize}
An AdamW optimizer is used with a fixed weight decay of $0.01$. Training incorporates early stopping with a patience of $10$ epochs. For object detection, the batch size is fixed at 8, and HPO is conducted solely on the learning rate. 
For datasets with multiple timestamps the following number of timestamps were used as the default unless otherwise stated: Biomassters and Pastis used 7 timestamps, Kuro Siwo used 2 timestamps (pre and post flood), Dynamic Earthnet used 6 timestamps ('weekly' setting). Additionally, the RMSE results obtained on Biomassters were converted into a positive metric (higher is better) with a utility function that can be found in the Leaderboard.
   
    \begin{figure*}[t]
    \centering
    \includegraphics[width=0.8\textwidth]{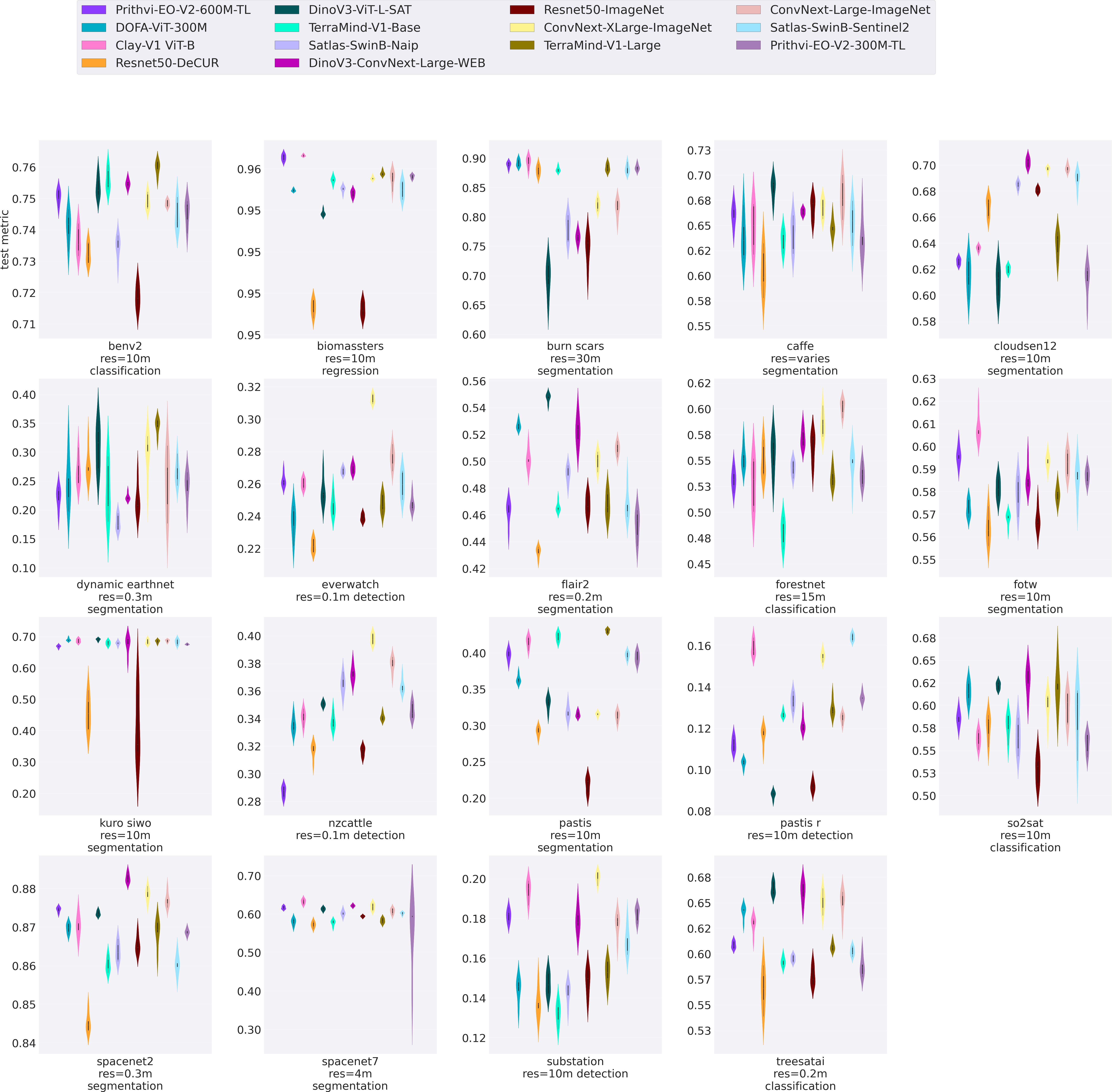}  
    \caption{Main results: Raw performance by dataset.}
    \label{fig:raw_results_per_dataset}
    \end{figure*}
    
\subsection{Additional Results}
\label{app:additional_results}
\subsubsection{Main Raw Results by Dataset}
\label{app:raw_results} 

The raw results obtained using these base experimental settings for each of the 19 datasets are presented in Figure \ref{fig:raw_results_per_dataset}. Beyond the main experimental settings, the succeeding sections give results from the various ablations conducted to understand the effect of changing settings on benchmarking.

\subsubsection{Frozen Encoder Ablation}
\label{app:frozen_results} 

\begin{figure}[b]
        \centering
        \includegraphics[width=1.0\linewidth]{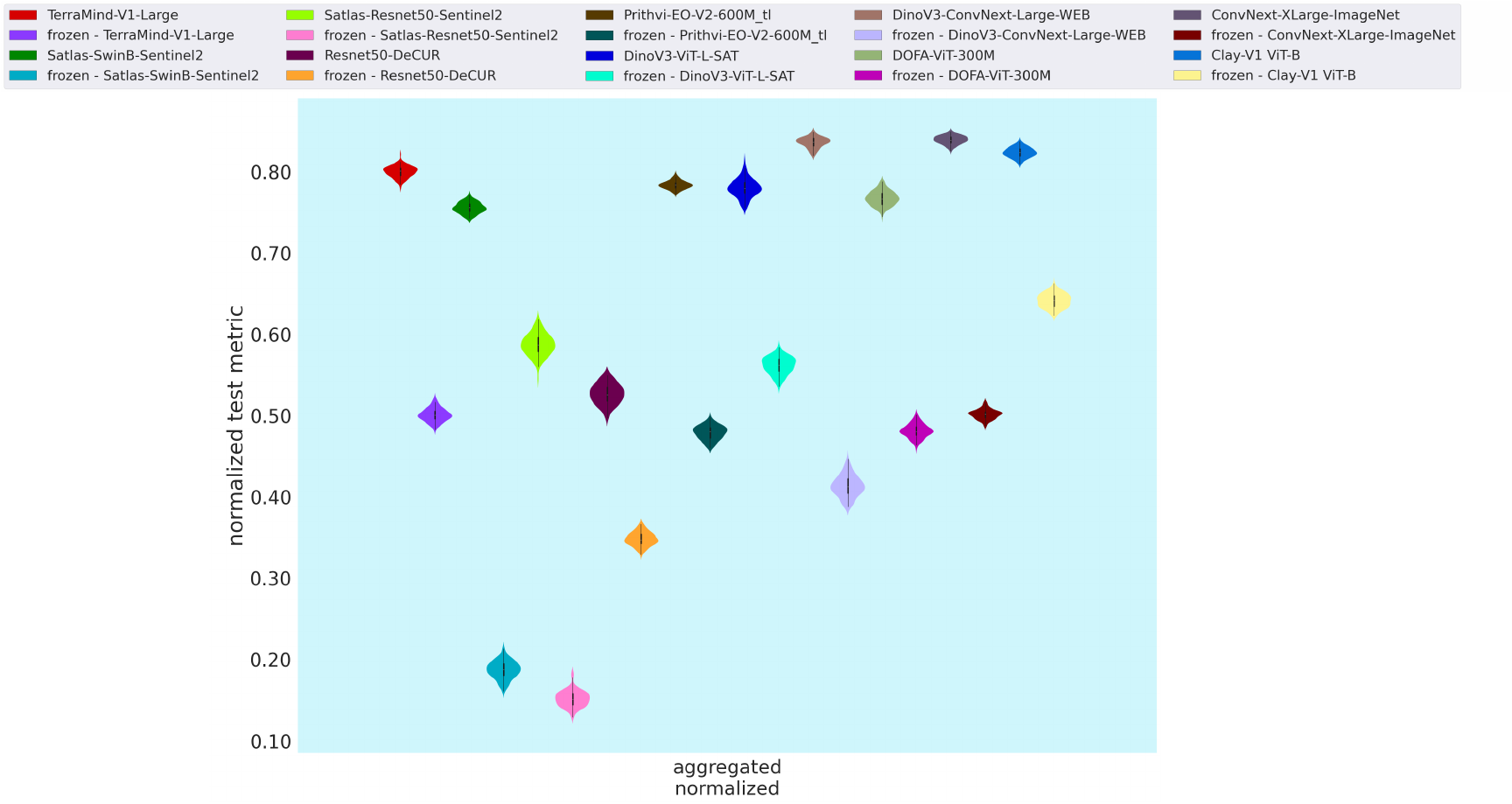}  
        \caption{Aggregated Frozen Vs Unfrozen Ablation: freezing the encoder results in a systematic drop in performance.}
        \label{fig:frozen_vs_main_aggregated}
    \end{figure}

\begin{figure*}[t]
        \centering
        \includegraphics[width=0.7\textwidth]{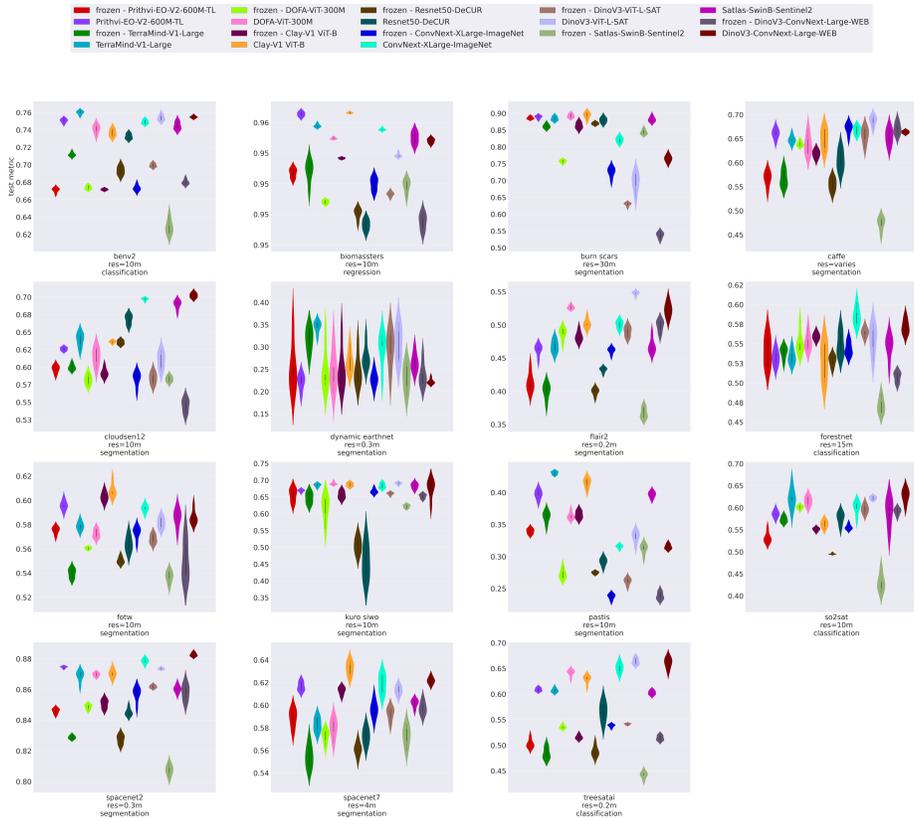}  
        \caption{Frozen Vs Unfrozen Encoder Ablation: Raw results per dataset.}
        \label{fig:frozen_vs_main_raw}
    \end{figure*}

To assess the impact of freezing the encoder during training, we performed an ablation where the encoder remains fixed while all other settings match the main experiments. This ablation was applied to all pixel-wise and classification datasets using a subset of models. Figure \ref{fig:frozen_vs_main_aggregated} shows a side-by-side comparison of full fine-tuning versus frozen encoder aggregated performance for each model, while Figure~\ref{fig:frozen_vs_main_raw} presents the raw results per dataset. A systematic drop in performance is observed when the encoder is frozen, suggesting that all models benefit from some fine-tuning. Importantly, a change in model ranking is also observed. Figure \ref{fig:rank_analysis_ablation} shows the aggregate change in model rank due to this ablation along with rank changes due to other ablations.

    \begin{figure}[t]
        \centering
        \includegraphics[width=0.55\linewidth]{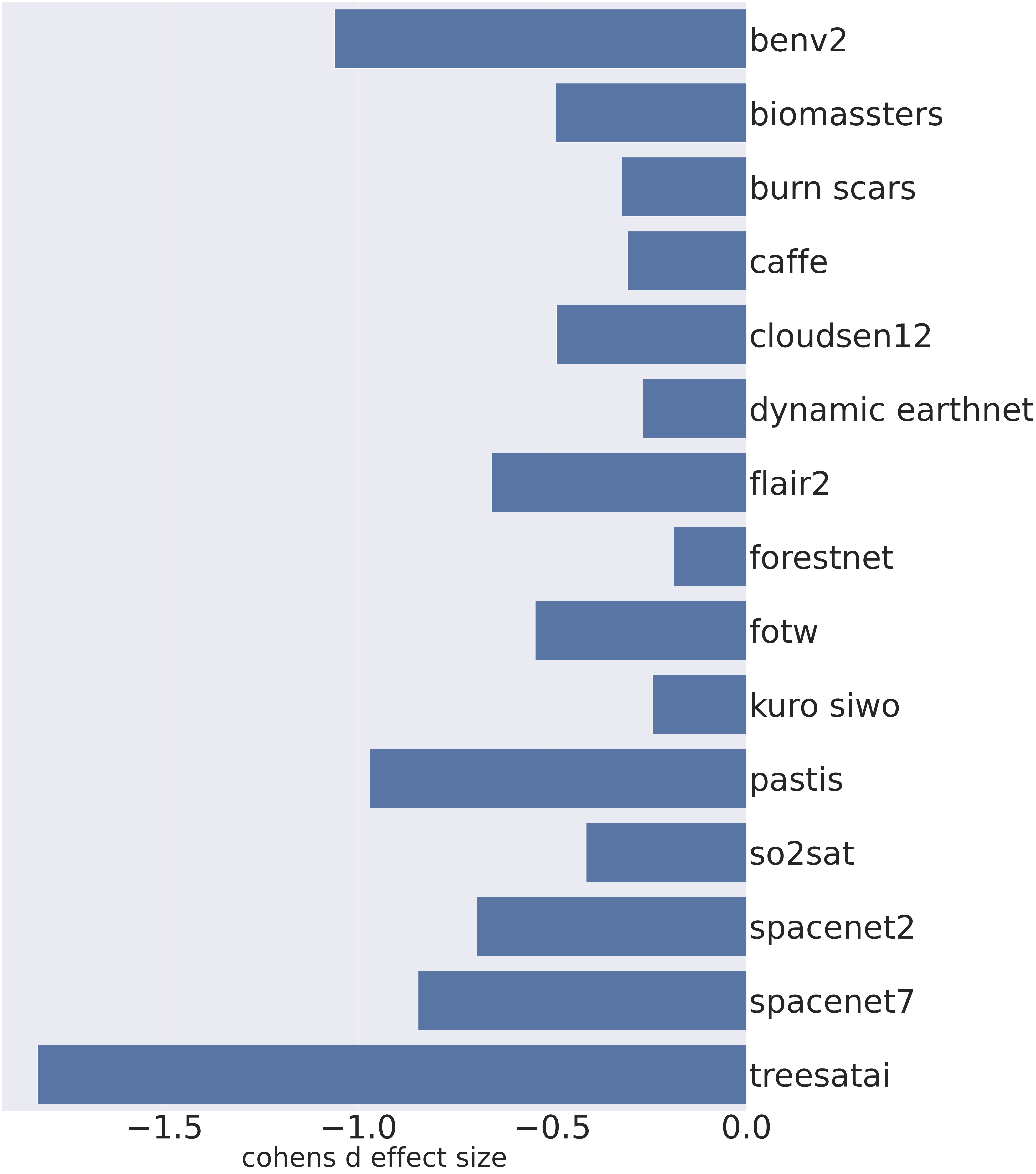}
        \caption{Cohen's D effect size for Unfrozen/Frozen Ablation: Per dataset reduction in performance due to freezing encoder }
        \label{fig:frozen_vs_main_cohens}
    \end{figure} 
    It should be noted that the drop in performance is not the same across all datasets. To investigate the effect size by dataset, a paired t-test is performed and the Cohen's D effect size is computed for datasets that have a statistically significant change. Figure \ref{fig:frozen_vs_main_cohens} shows the Cohen's D effect size for datasets with a statistically significant drop in test metric due to frozen backbones (aggregated across all models). Datasets such as treesatai and benv2 have the biggest drop in performance with encoder freezing.




\subsubsection{Linear Decoder Ablation}
\label{app:linear_results} 
To evaluate the impact of using a linear decoder for pixel-wise tasks, we ran an ablation replacing the UNet decoder with a single linear layer, keeping all other settings identical to the main experiments. This was applied to all pixel-wise datasets (segmentation and regression) using a subset of models. 
    \begin{figure}[b]
        \centering
        \includegraphics[width=1.0\linewidth]{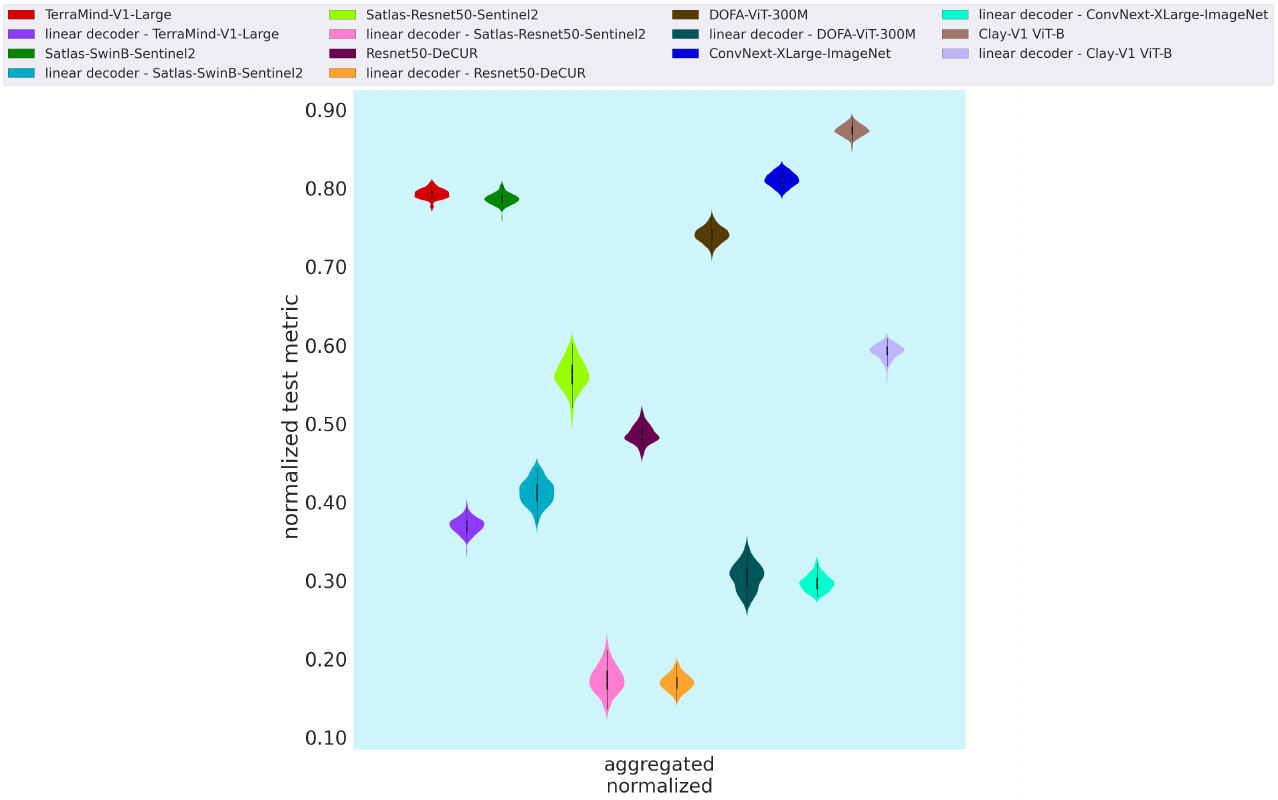}  
        \caption{Aggregated UNet Vs Linear Decoder Ablation: linear decoder lowers performance for all models.}
        \label{fig:linear_vs_main_aggregated}
    \end{figure}

    \begin{figure*}[t]
        \centering
        \includegraphics[width=0.6\textwidth]{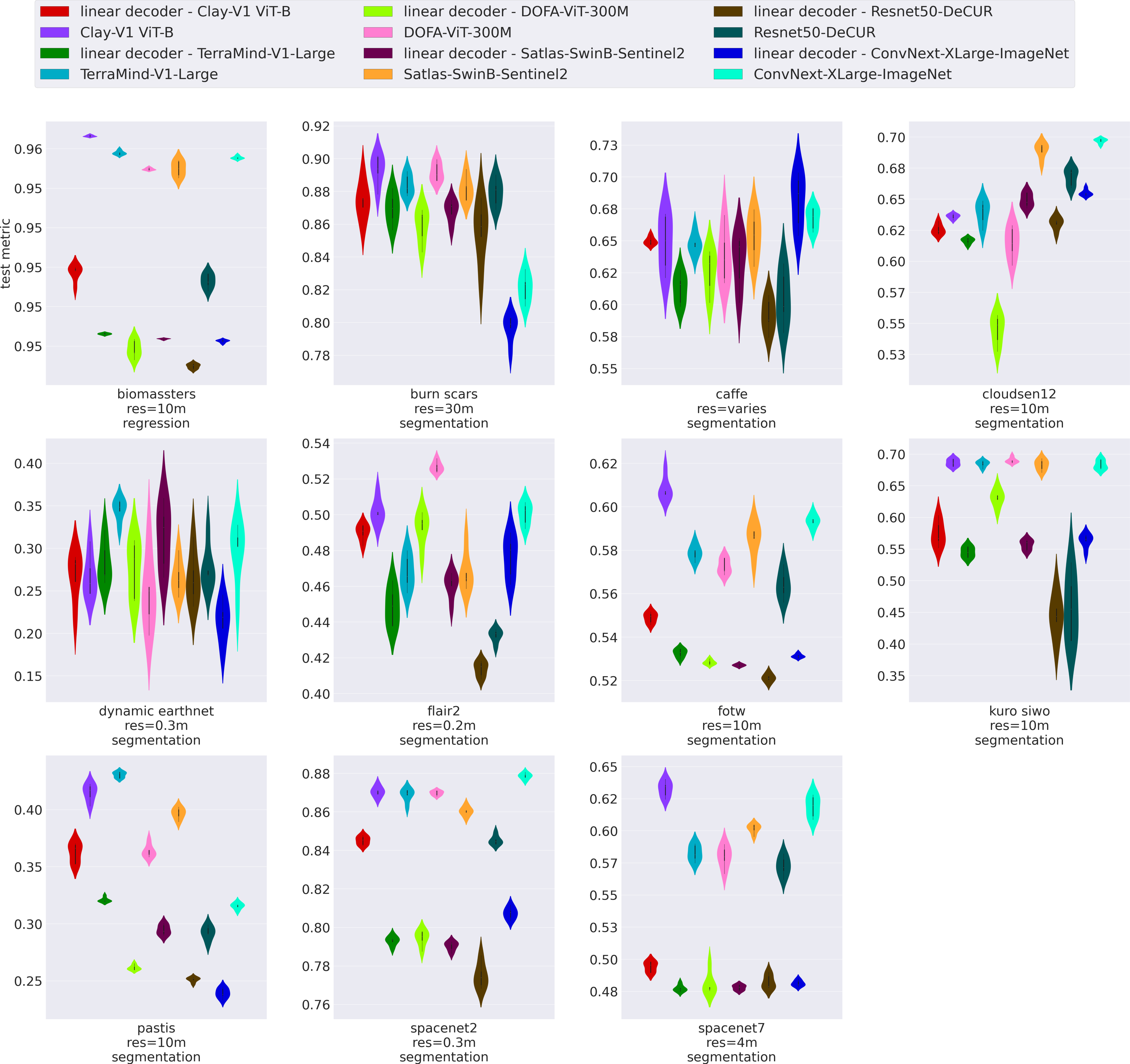}  
        \caption{UNet Vs Linear Decoder Ablation: Raw results per dataset.}
        \label{fig:linear_vs_main_raw}
    \end{figure*}

\begin{figure}[t]
            \centering
            \includegraphics[width=0.55\linewidth]{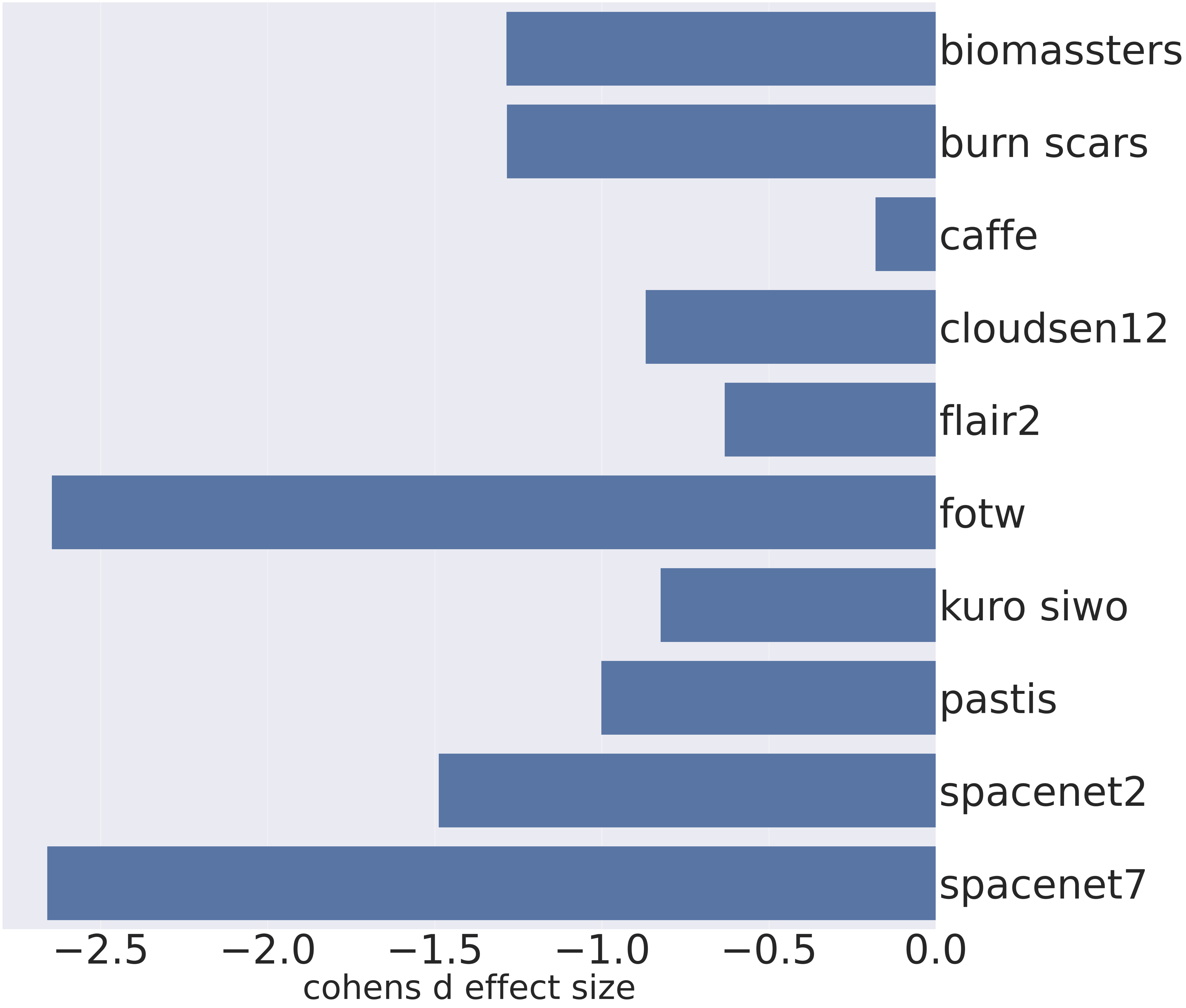}
            \caption{Cohen's D effect size for UNet/Linear Decoder Ablation: Per dataset reduction in performance due to linear decoder.}
            \label{fig:linear_vs_main_cohens}
        \end{figure}

Figure~\ref{fig:linear_vs_main_aggregated} compares aggregated performance of UNet versus Linear decoder for each model, while Figure~\ref{fig:linear_vs_main_raw} shows the raw results per dataset. Again, a linear decoder shows lower performance and substantially affects scoring. Figure  \ref{fig:linear_vs_main_cohens} shows variation in Cohen's D effect size per dataset for datasets with statistically significant reduction. In particular, fotw and spacenet7 datasets have the largest decrease. Figure \ref{fig:rank_analysis_ablation} demonstrates that while there is an aggregate change in model rank due to this ablation, the difference in rank is smaller than changes observed due to other ablations.

\subsubsection{Multi-Spectral Vs RGB Data Ablation}
\label{app:rgb_results} 

To assess the role of multi-spectral data, we ran an ablation using only GeoFM encoders on datasets with additional bands beyond RGB. Unlike the main experiments, which used all available bands, this ablation restricted inputs to RGB only, while keeping all other settings unchanged. 
    \begin{figure}[b]
        \centering
        \includegraphics[width=1.0\linewidth]{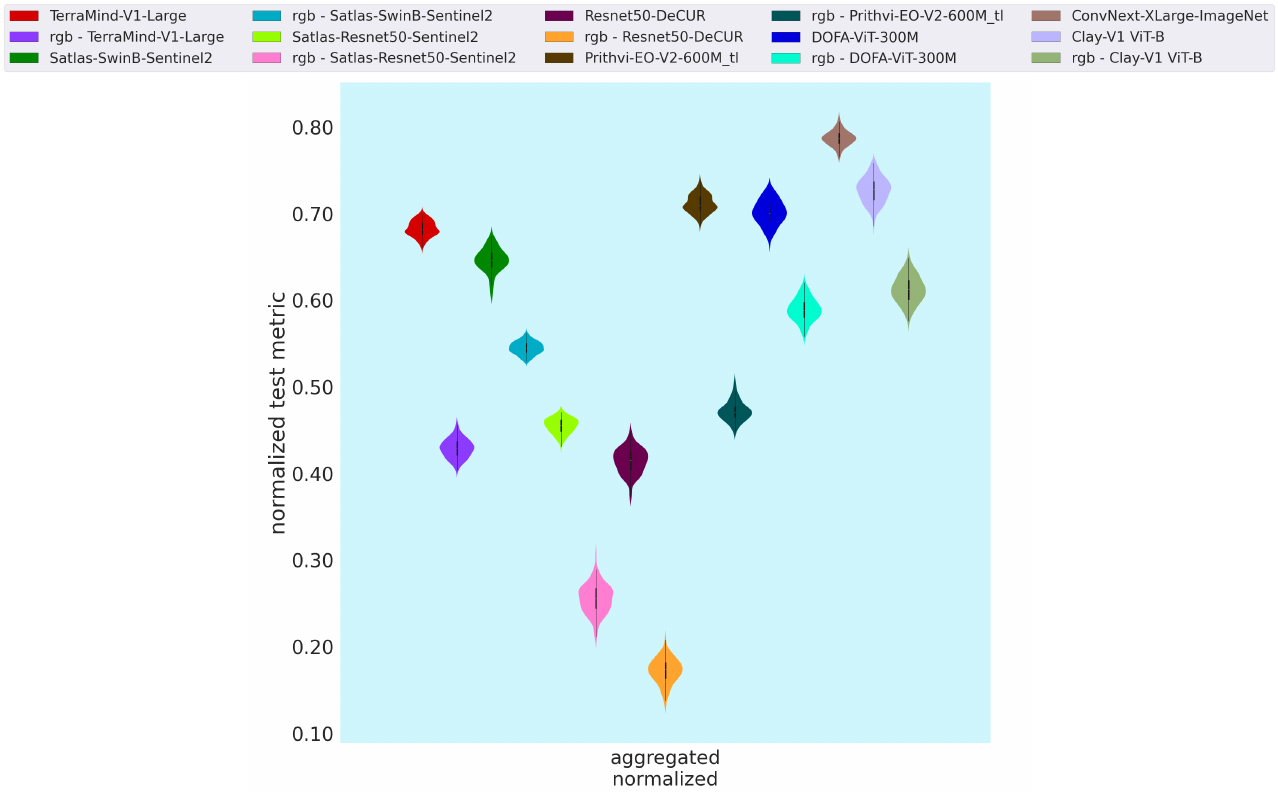}  
        \caption{Aggregated Multi-Spectral Vs RGB Data Ablation: excluding multi-spectral data tends to reduce performance.}
        \label{fig:rgb_vs_main_aggregated}
    \end{figure}

    \begin{figure*}[t]
        \centering
        \includegraphics[width=0.6\textwidth]{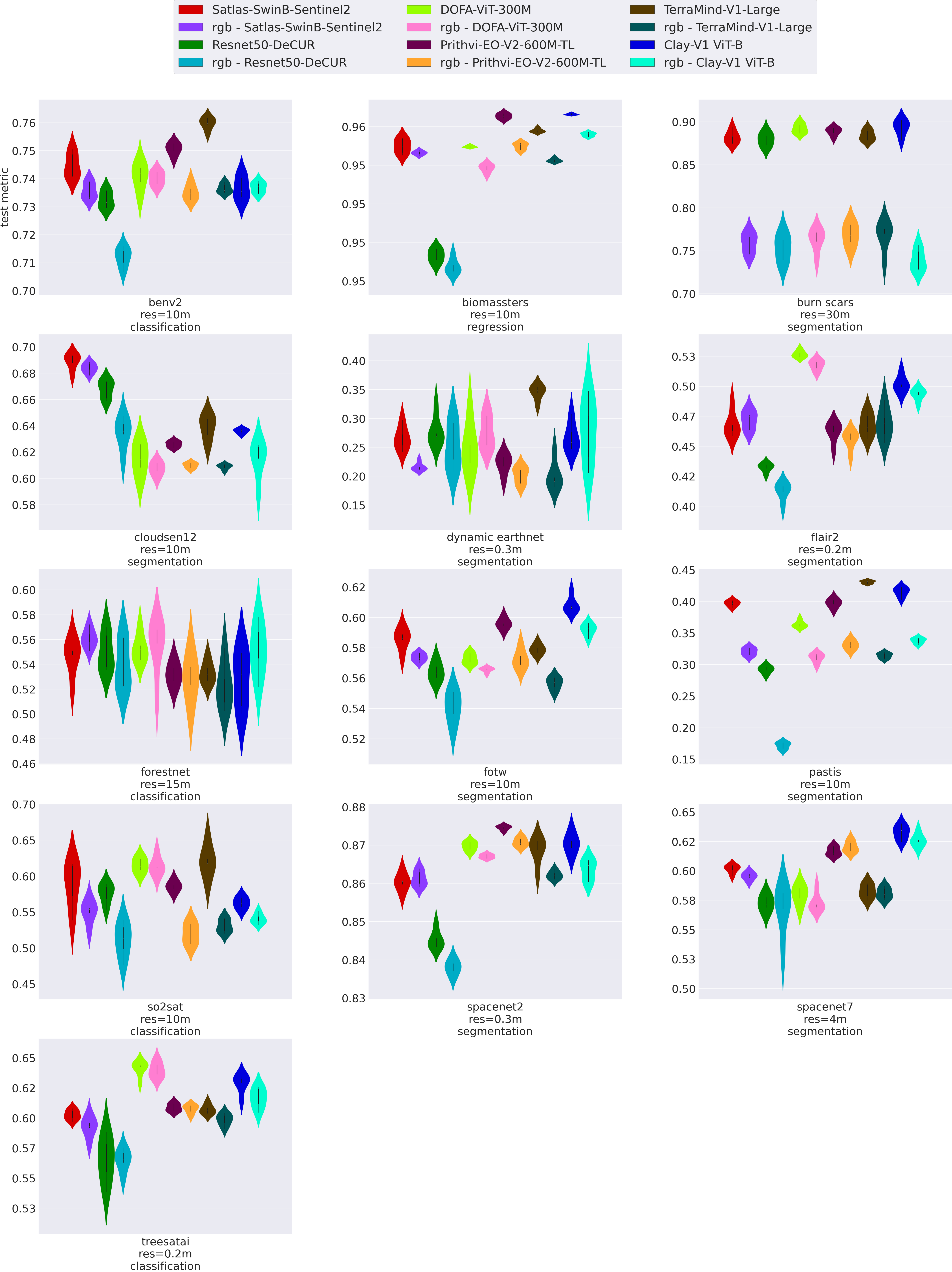}  
        \caption{Multi-Spectral Vs RGB Data Ablation: Raw results per dataset.}
        \label{fig:rgb_vs_main_raw}
    \end{figure*}

    \begin{figure}[t]
        \centering
        \includegraphics[width=0.55\linewidth]{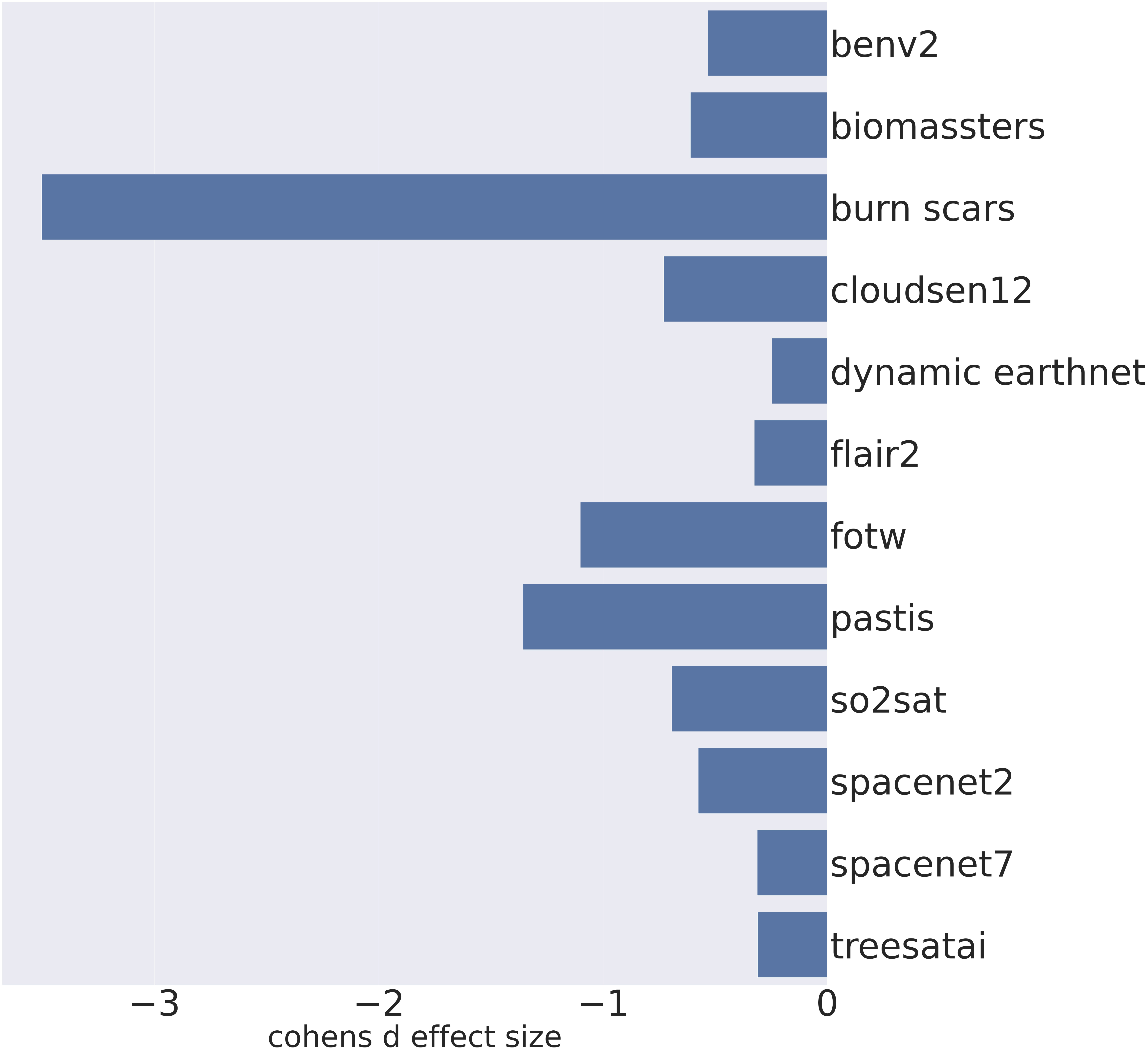}
        \caption{Cohen's D effect size for Multi-Spectral/RGB Ablation: Per dataset reduction in performance due to excluding multi-spectral data.}
        \label{fig:rgb_vs_main_cohens}
    \end{figure} 
    
Figure~\ref{fig:rgb_vs_main_raw} shows the raw per-dataset results, while \ref{fig:rgb_vs_main_aggregated} shows aggregated performance. A general drop in performance is observed in both figures when multi-spectral bands are excluded, thus highlighting the importance of multi-spectral data. 
Similar to the other ablations, the Cohen’s D effect size is computed for datasets that have a statistically significant change according to a paired t-test. Datasets with a statistically significant drop in performance due to excluding multi-spectral bands (aggregated across all models) are highlighted in Figure \ref{fig:rgb_vs_main_cohens}. The burn scar dataset suffers the biggest reduction by far, possibly due to some burn scars not being detectable with simple RGB data. This emphasizes the importance of multi-spectral data for some downstream tasks in EO. The Multi-Spectral-dependent capability is comprised of datasets with a statistically significant change.

\subsubsection{Ground Sampling Distance (GSD) Ablation}
\label{app:gsd_results} 
This ablation applies a 14x14 kernel filter to increase the Ground Sampling Distance (GSD) of validation and test images, while train images are kept at the original resolution. The best HPO settings are used for training with 5 repeated seeds for each model on each dataset. 
    \begin{figure*}[t]
        \centering
        \includegraphics[width=0.7\textwidth]{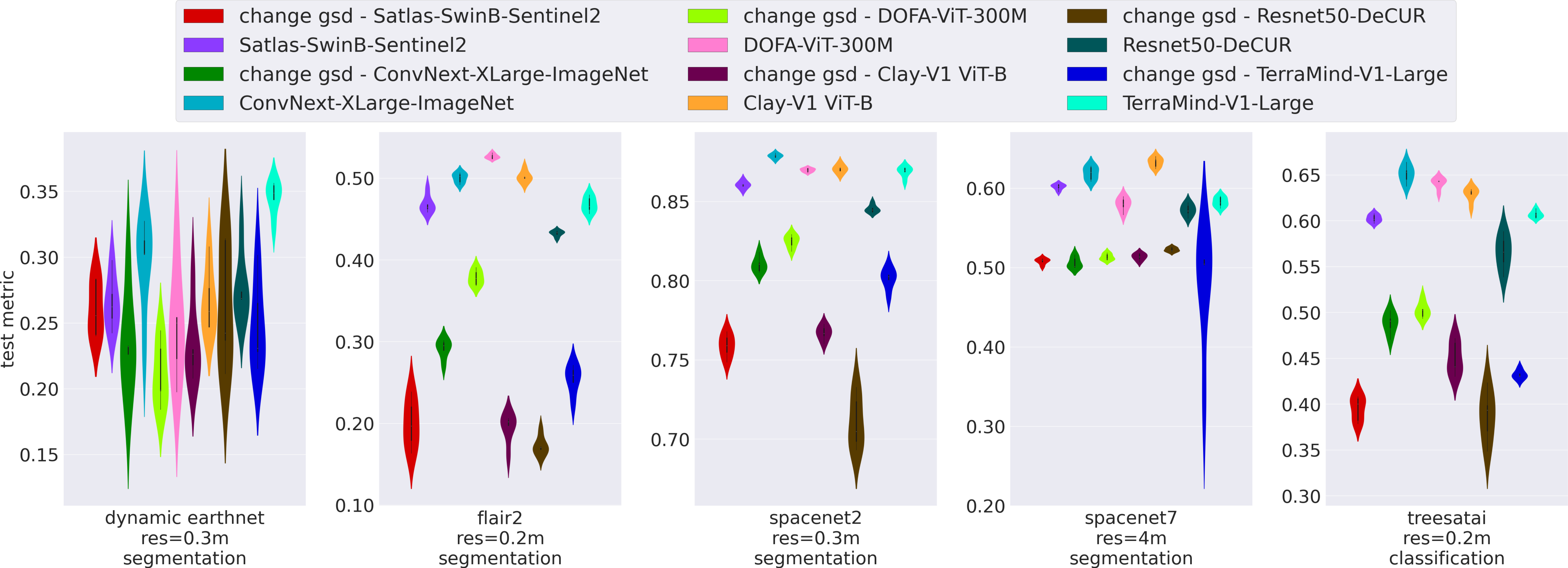}  
        \caption{Increasing Ground Sampling Distance Ablation: Raw results on 5 datasets.}
        \label{fig:change_gsd_vs_main_raw}
    \end{figure*}
    
    \begin{figure}[t]
        \centering
        \includegraphics[width=1.0\linewidth]{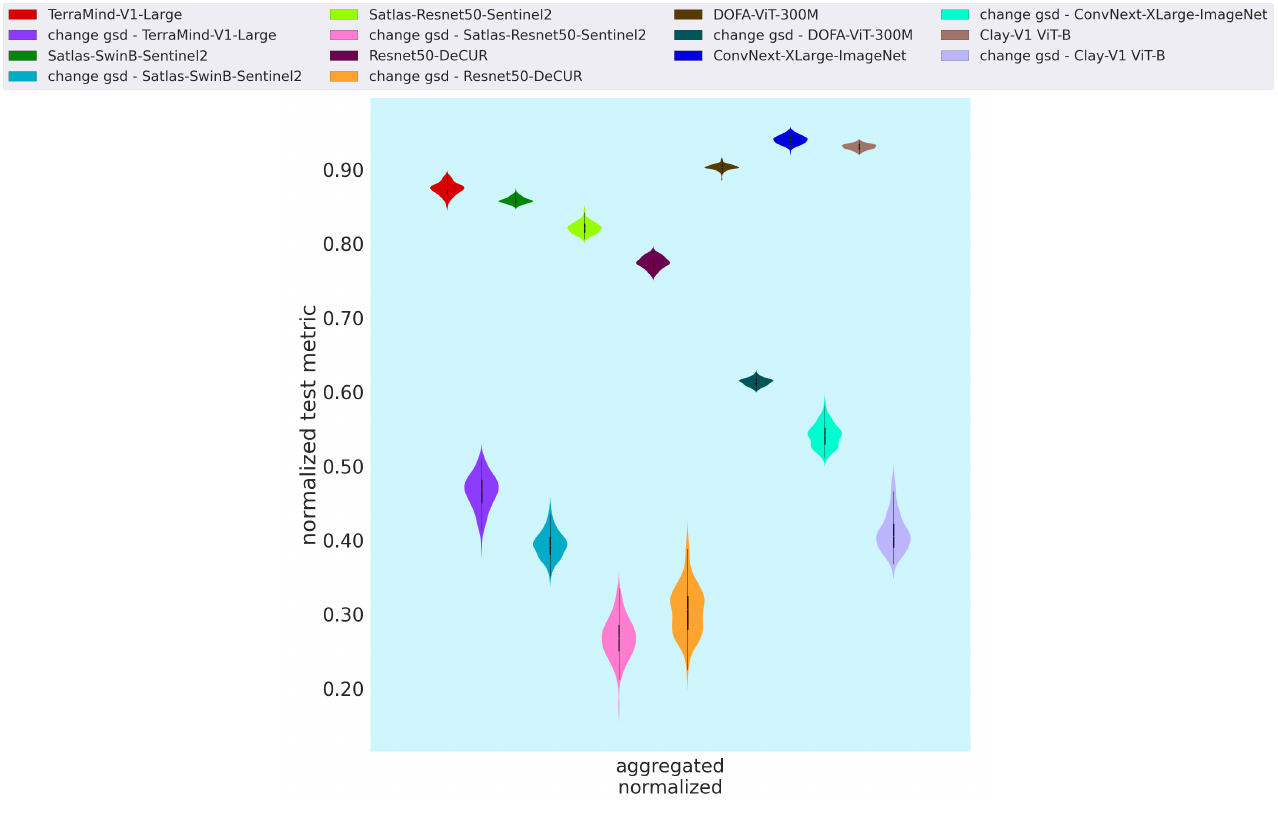}  
        \caption{Aggregated Original GSD Vs Increased GSD: lower GSD can improve performance in some cases.}
        \label{fig:change_gsd_vs_main_aggregated}
    \end{figure}

    \begin{figure}[t]
        \centering
        \includegraphics[width=0.55\linewidth]{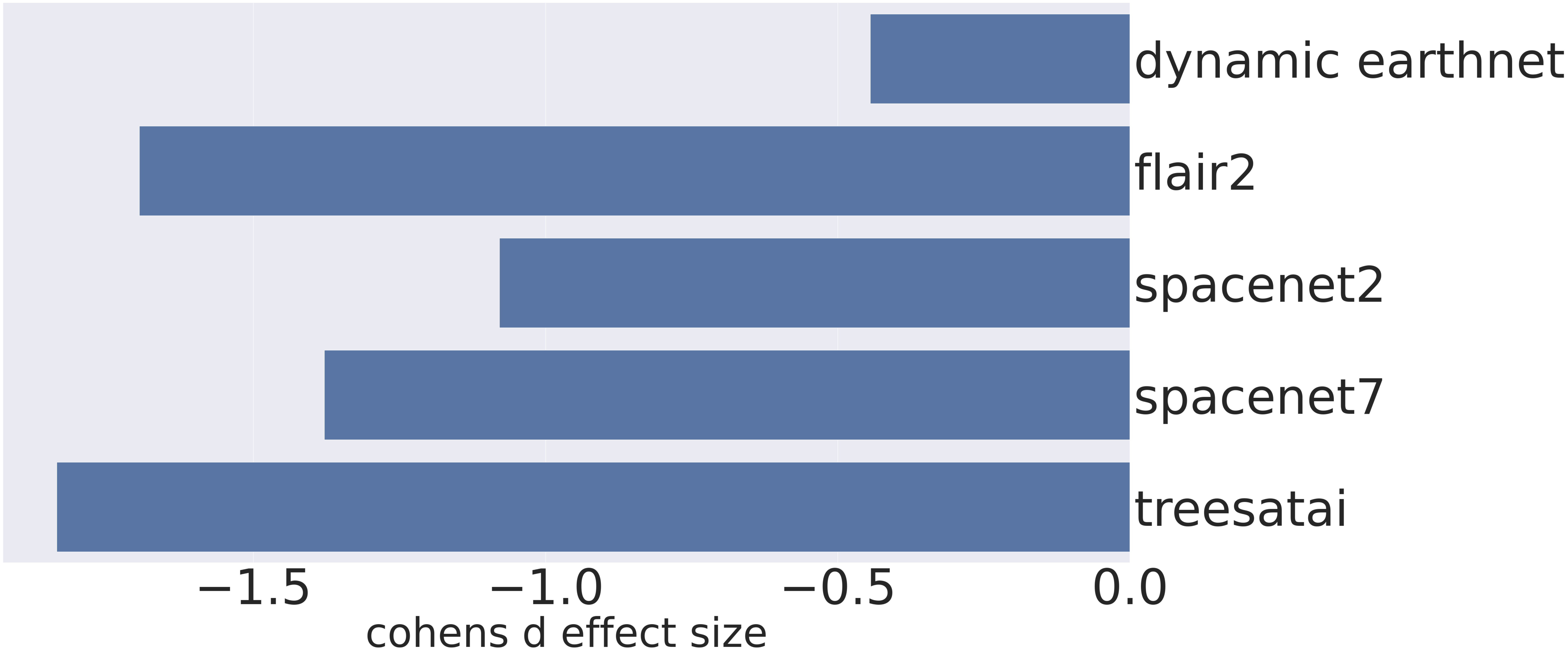}
        \caption{Cohen's D effect size for Increasing Ground Sampling Distance Ablation: Per dataset reduction in performance due to increased GSD.}
        \label{fig:change_gsd_vs_main_cohens}
    \end{figure}

All other settings used in the main experiments were maintained. Datasets with a resolution under 10m are used with a subset of models. Ground sampling distance is an important factor in satellite data, determining the level of detail that can be observed in a satellite image. A lower GSD (high resolution) implies that each pixel covers a smaller area on the ground and finer details are clearer when looking at the overall image. Figure \ref{fig:change_gsd_vs_main_raw} shows raw results per dataset from this ablation, while \ref{fig:change_gsd_vs_main_aggregated} shows aggregated performance. Both figures highlight that increasing GSD tends to lower test metrics across all models.
 This is expected as the images with lower (original) GSD have more details. However, it is notable that some datasets exhibit a smaller effect size as demonstrated by the results Cohen's D effect sizes of Figure \ref{fig:change_gsd_vs_main_cohens}.

\subsubsection{Multi-Modality Ablation}
\label{app:multimodal_results} 
\begin{figure}[t]
        \centering
        \includegraphics[width=0.9\linewidth]{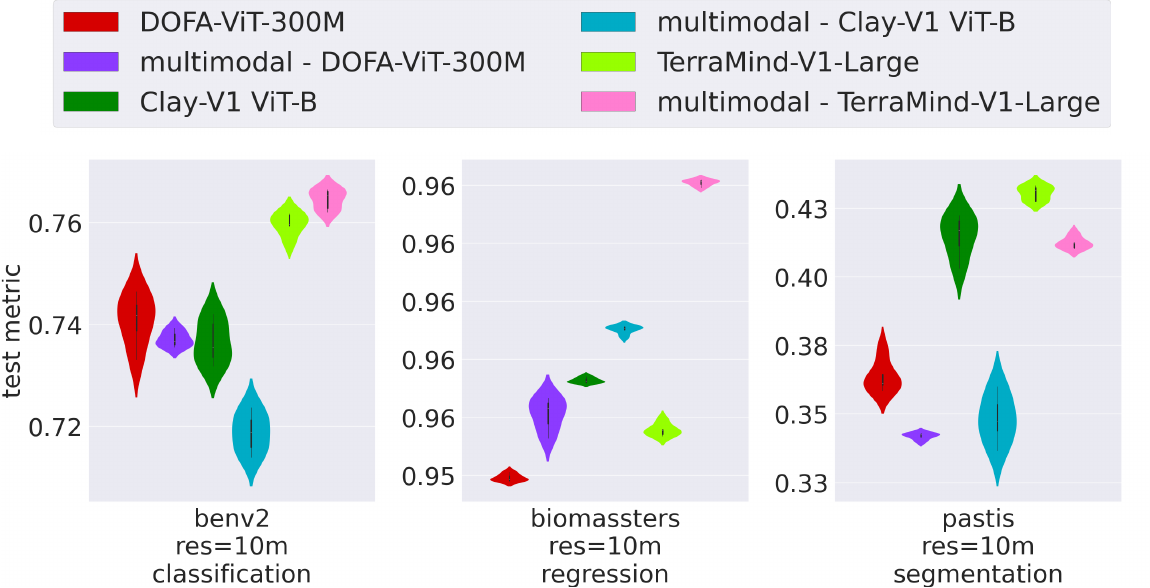}  
        \caption{Multi-modal Vs Optical only: change in performance with multi-modal input vs Optical input only. Ablation on all multi-modal datasets and models only}
        \label{fig:multi_modal_raw}
    \end{figure}

To investigate the effect of multi-modal inputs, this ablation applies both Optical and SAR data simultaneously to each model versus only using optical data (e.g. Sentinel 2 or Planet data).  All other settings used in the main experiments were maintained. The raw results of Figure \ref{fig:multi_modal_raw} show that this ablation is particularly dependent on the dataset and model.

For the pastis dataset, multi-modal inputs reduce test metrics regardless of the model, suggesting that the combination of the multi-modal inputs may not be beneficial for this dataset. Conversely, the biomassters dataset consistently shows an improvement across models. Benv2 dataset has more nuanced results: while TerraMind-V1-Large increases test metrics with multi-modal inputs, the remaining models have either similar or lower performance. Overall, the effect of multi-modal inputs is inconclusive as performance appears to vary by model and dataset. 




\subsubsection{Multi-Temporal Vs Mono-Temporal Ablation}
\label{app:multitemp_results} 
To evaluate the effect of using a single timestamp, we ran an ablation on multi-temporal datasets where inputs were restricted to one timestamp instead of multiple (up to seven in the main experiments), keeping all other settings unchanged. 

    \begin{figure*}[t]
    \centering
    \includegraphics[width=0.6\textwidth]{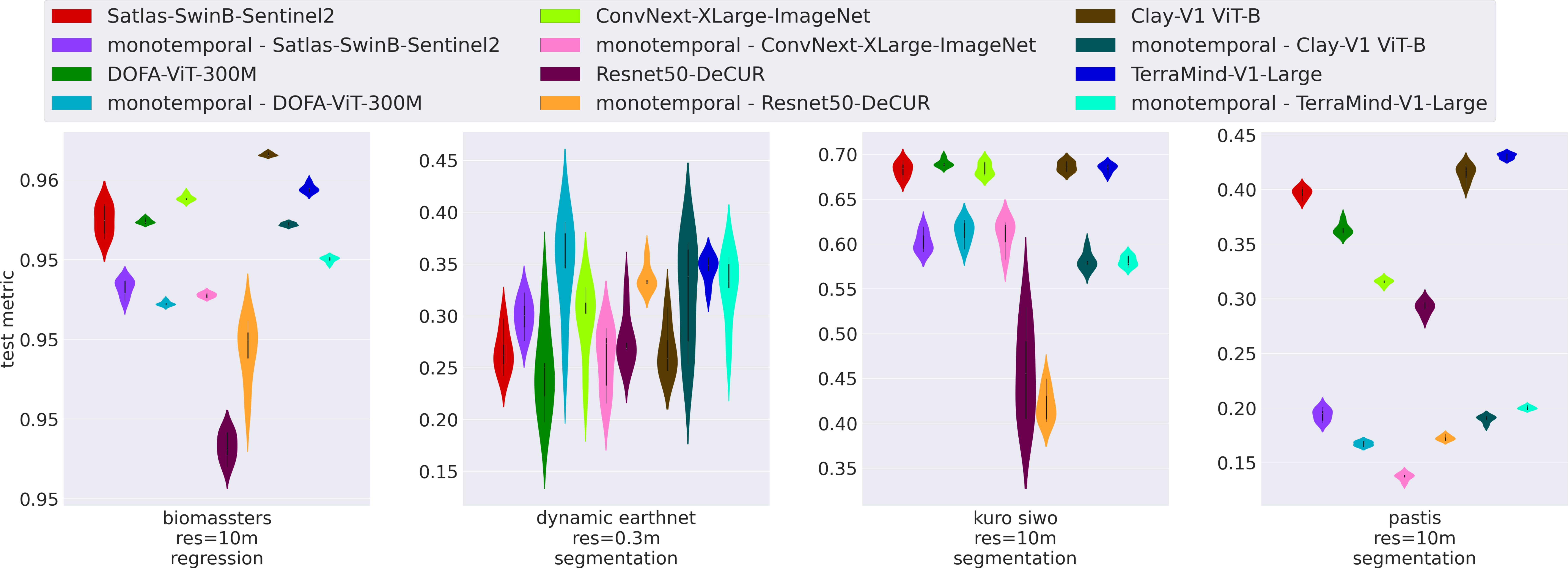}  
    \caption{Multi-Temporal Vs Mono-Temporal Ablation: raw results on 4 datasets.}
    \label{fig:monotemp_vs_main_raw}
    \end{figure*}

 Applied to a subset of models, this ablation revealed substantial performance drops across some datasets such as pastis, thus confirming the importance of temporal information for these particular datasets in the multi-temporal capability. Raw per-dataset results are shown in Figure~\ref{fig:monotemp_vs_main_raw}. Additionally, Figure \ref{fig:rank_analysis_ablation} also shows that changing from multi-temporal to mono-temporal inputs causes the biggest change in rank, second only to the effect of freezing the encoder.


    
    

\subsubsection{Cosine Annealing Ablation}
\label{app:cosine_results} 
\begin{figure}[t]
    \centering
    \includegraphics[width=0.9\linewidth]{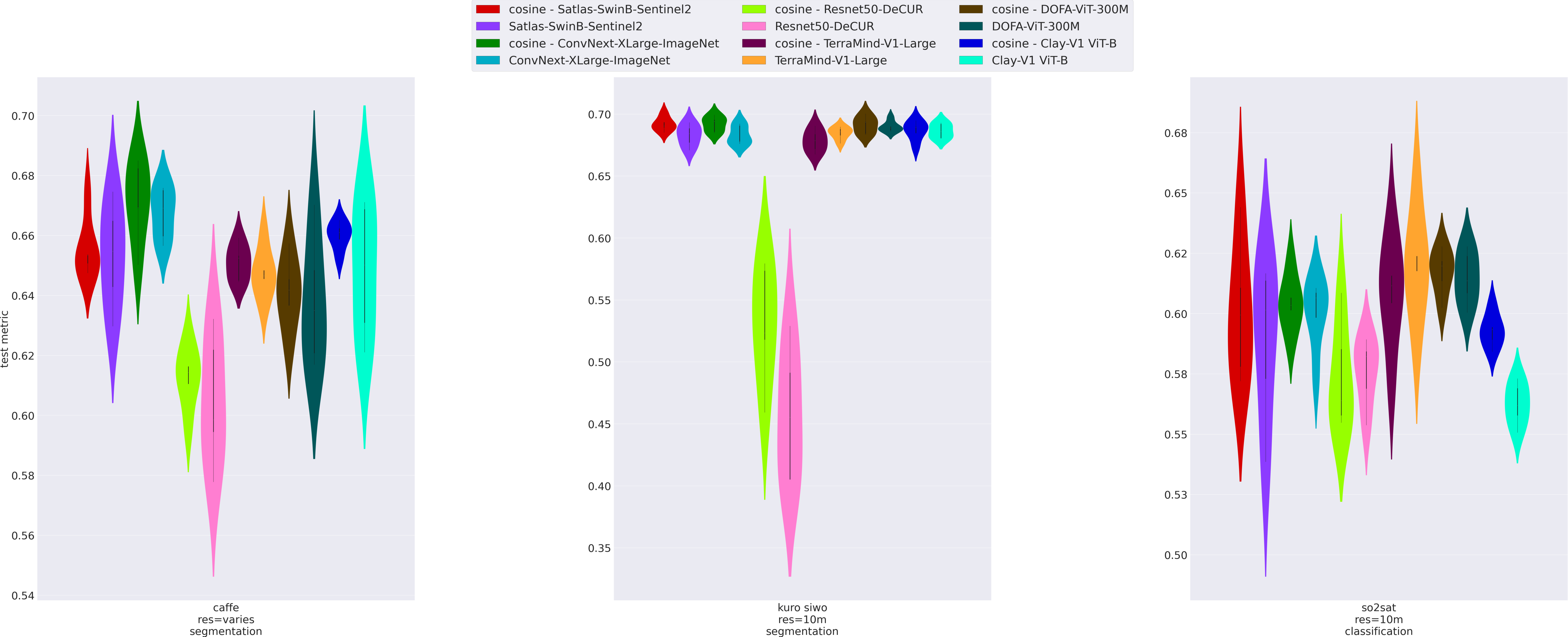}  
    \caption{Cosine annealing Learning rate Vs Fixed Learning rate: raw results on 3 datasets are inconclusive.}
    \label{fig:cosine_vs_main_raw}
    \end{figure}

In this ablation, we applied a cosine annealing schedule to adjust the learning rate during training. Figure~\ref{fig:cosine_vs_main_raw} shows the raw per-dataset results. Outcomes varied across datasets and models, providing no definitive conclusion.



\subsubsection{Linear Warm Up  Ablation}
\label{app:linear_warm_up_results} 
\begin{figure}[t]
    \centering
    \includegraphics[width=0.9\linewidth]{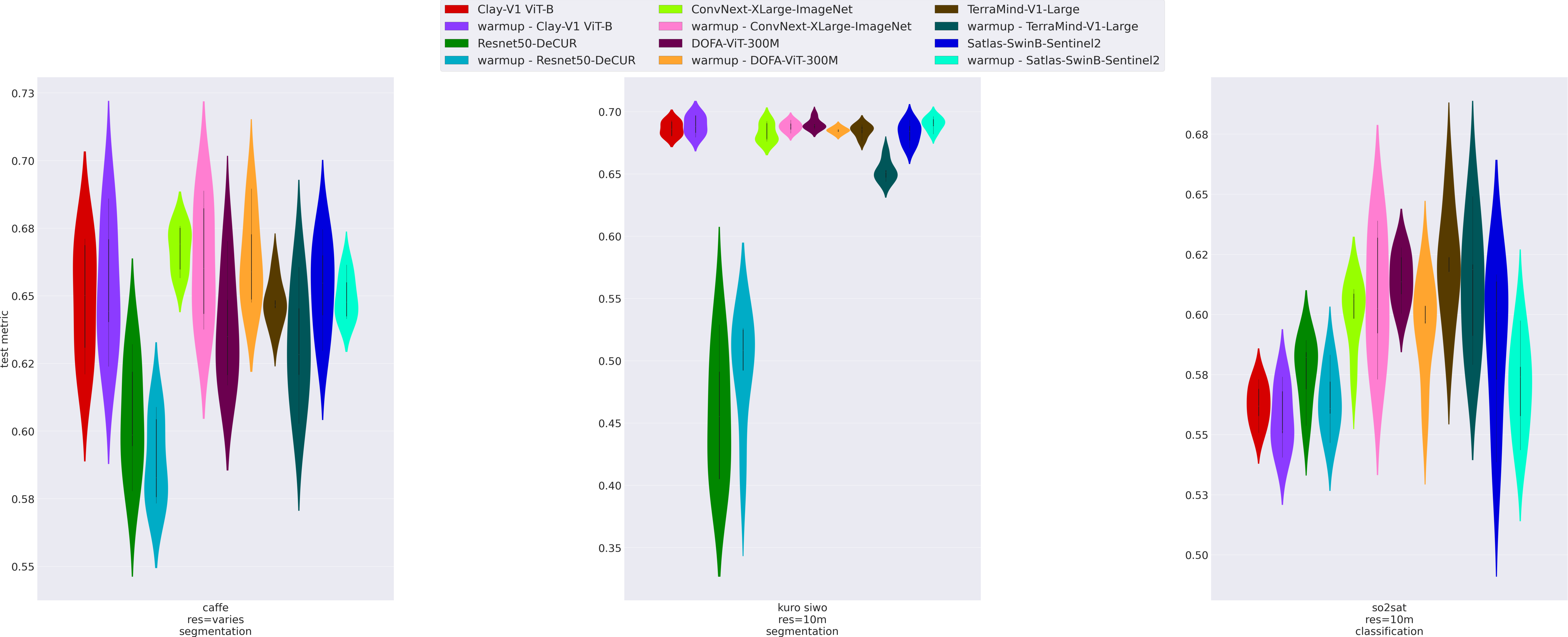}  
    \caption{Linear Warm Up on Learning rate Vs Fixed Learning rate: raw results on 3 datasets are inconclusive.}
    \label{fig:warmup_vs_main_raw}
    \end{figure}

In this ablation, we applied linear warm-up to adjust the learning rate during the first five training epochs. Figure~\ref{fig:warmup_vs_main_raw} shows the raw per-dataset results. Outcomes varied across datasets and models not showing any clear trend.

\end{document}